\documentclass[12pt]{article}
\usepackage[left]{lineno}
\usepackage{placeins}  
\usepackage{cite}
\usepackage{times}
\usepackage{caption}
\usepackage{subcaption}
\usepackage[table,xcdraw]{xcolor}
\usepackage{booktabs}
\usepackage{siunitx}
\usepackage{todonotes}
\usepackage[normalem]{ulem}
\usepackage{graphicx}
\usepackage{amsmath,amssymb,amsfonts}
\graphicspath{ {figures/} }
\usepackage{parskip}
\usepackage[misc]{ifsym}
\usepackage[pdfborder={0 0 0},pdftex,hidelinks]{hyperref}
\usepackage{upgreek}

\usepackage{microtype}
\usepackage{booktabs} % for professional tables
\usepackage{xspace}
\usepackage{titletoc}
\usepackage{enumitem}
\usepackage{xfrac}
\usepackage{wrapfig}
\usepackage{bm}
\usepackage{algorithm, algorithmic, caption}
\usepackage{dsfont}
\usepackage{pifont}
\usepackage{adjustbox}
\usepackage[dvipsnames]{xcolor}
\usepackage{multirow}
\usepackage{array}    % for !{\vrule}
\usepackage{makecell} % for \Xhline

\definecolor{Gold}{rgb}{1,0.84,0}
\definecolor{Bronze}{rgb}{0.69,0.55,0.34}

\definecolor{chaptercolor}{HTML}{1A254B}
\definecolor{darkblue}{HTML}{1A254B}
\definecolor{linkcolor}{HTML}{2B50AA}
\definecolor{citecolor}{HTML}{2B50AA}
\definecolor{lightlinkcolor}{HTML}{9A8F97}
\definecolor{darklinkcolor}{HTML}{1A254B}
\definecolor{pink}{HTML}{E05F60}
\definecolor{lightblue}{HTML}{A7BED3}
\definecolor{red}{HTML}{F2545B}
\definecolor{blue}{HTML}{2b50aa}

\usepackage{amsmath}
\usepackage{amssymb}
\usepackage{mathtools}
\usepackage{amsthm}
\usepackage{bbm}
\usepackage{nicefrac}
\usepackage[capitalize,noabbrev]{cleveref}

\usepackage[para,online,flushleft]{threeparttable}
\usepackage{adjustbox}
\theoremstyle{plain}

\theoremstyle{definition}

\theoremstyle{remark}

\Crefname{assumption}{Assumption}{Assumptions}
\crefname{assumption}{Assumption}{Assumptions}
\allowdisplaybreaks
\makeatletter
\renewcommand{\paragraph}{%
  \@startsection{paragraph}{4}%
  {\z@}{0ex \@plus 0ex \@minus 0ex}{-1em}%
  {\normalfont\normalsize\bfseries}%\
}
\makeatother

\newlist{lemenum}{enumerate}{1} % should only occur inside lem env.
\setlist[lemenum]{label=(\roman*),ref=\thelemma\,(\roman*),topsep=0pt}
\Crefname{lemenumi}{Lemma}{Lemmas}

\newlist{corenum}{enumerate}{1} % should only occur inside cor env.
\setlist[corenum]{label=(\roman*),ref=\thecorollary\,(\roman*),topsep=0pt}
\Crefname{corenumi}{Corollary}{Corollaries}

\newlist{thmenum}{enumerate}{1} % should only occur inside thm env.
\setlist[thmenum]{label=(\roman*),ref=\thetheorem\,(\roman*),topsep=0pt}
\Crefname{thmenumi}{Theorem}{Theorems}

\newlist{propenum}{enumerate}{1} % should only occur inside prop env.
\setlist[propenum]{label=(\roman*),ref=\thedefinition\,(\roman*),topsep=0pt}
\Crefname{propenumi}{Property}{Properties}

\newlist{assenum}{enumerate}{1} % should only occur inside assumption env.
\setlist[assenum]{label=(\roman*),ref=\theassumption\,(\roman*),topsep=0pt}
\Crefname{assenumi}{Assumption}{Assumptions}

\usepackage{svg}
\usepackage{graphicx}
\usepackage{rotating}
\usepackage{tikz}
\usepackage{pgfplots}
\pgfplotsset{compat=newest}
\usetikzlibrary{shapes.geometric}

\usepackage{import}
\usepackage{xifthen}
\usepackage{pdfpages}
\usepackage{transparent}

\NewDocumentCommand{\incfig}{mo}{
  \begin{center}
    \IfValueT{#2}{\def\svgwidth{#2}}{\def\svgwidth{\columnwidth}}
    \import{./figures/}{#1.pdf_tex}
  \end{center}
}

\usepackage{pgf}
\usepackage{adjustbox}

\NewDocumentCommand{\incplt}{O{\columnwidth}m}{%
  \begin{center}
    \adjustbox{width=#1}{\import{./plots/output/}{#2.pgf}}
  \end{center}
}



\usepackage{aligned-overset}

\DeclareFontFamily{U}{mathb}{\hyphenchar\font45}
\DeclareFontShape{U}{mathb}{m}{n}{
      <5> <6> <7> <8> <9> <10> gen * mathb
      <10.95> mathb10 <12> <14.4> <17.28> <20.74> <24.88> mathb12
      }{}
\DeclareSymbolFont{mathb}{U}{mathb}{m}{n}
\DeclareFontSubstitution{U}{mathb}{m}{n}
\DeclareMathSymbol{\Asterisk}      {2}{mathb}{"06}
\newcommand{\mainsym}{\textcolor{blue}{\parentheses*{\hspace{-0.1em}\Asterisk\hspace{-0.1em}}}}

\newcommand{\psym}{\ensuremath{\textcolor{blue}{\dagger}}\xspace}

\NewDocumentCommand{\norm}{sm}{\IfBooleanTF{#1}{\|#2\|}{\left\| #2 \right\|}}
\newcommand{\setmath}[1]{\left\{#1\right\}}

\DeclareMathOperator*{\argmax}{arg\,max}

\DeclarePairedDelimiter\parentheses{(}{)}
\DeclarePairedDelimiter\brackets{[}{]}

\newcommand{\R}{\mathbb{R}}
\newcommand{\E}{\mathbb{E}}

\renewcommand{\vec}[1]{{\bm{#1}}}
\newcommand{\mat}[1]{\bm{#1}}

\NewDocumentCommand{\fnPr}{}{\mathbb{P}}
\RenewDocumentCommand{\Pr}{om}{\fnPr\IfValueT{#1}{_{#1}}\parentheses*{#2}}
\NewDocumentCommand{\Prsm}{om}{\fnPr\IfValueT{#1}{_{#1}}\parentheses{#2}}
\RenewDocumentCommand{\H}{mo}{\mathrm{H}\IfValueTF{#2}{\!\left[#1\ \middle|\ #2\right]}{\brackets*{#1}}}
\NewDocumentCommand{\Hsm}{mo}{\mathrm{H}\IfValueTF{#2}{[#1 \mid #2]}{\brackets{#1}}}
\NewDocumentCommand{\I}{mmo}{\mathrm{I}\IfValueTF{#3}{\!\left(#1;#2\ \middle|\ #3\right)}{\parentheses*{#1; #2}}}
\NewDocumentCommand{\Ism}{mmo}{\mathrm{I}\IfValueTF{#3}{(#1;#2 \mid #3)}{\parentheses{#1; #2}}}

\NewDocumentCommand{\ExpVal}{somo}{\ensuremath{\mathbb{E}\IfValueT{#2}{_{#2}}{} \IfBooleanTF{#1}{#3}{\IfValueTF{#4}{\!\left[#3\ \middle|\ #4\right]}{\brackets*{#3}}}}}
\NewDocumentCommand{\Esm}{somo}{\ensuremath{\mathbb{E}\IfValueT{#2}{_{#2}}{} \IfBooleanTF{#1}{#3}{\IfValueTF{#4}{\!\left[#3\ \middle|\ #4\right]}{\brackets{#3}}}}}
\NewDocumentCommand{\Var}{somo}{\mathrm{Var}\IfValueT{#2}{_{#2}}{} \IfBooleanTF{#1}{#3}{\IfValueTF{#4}{\!\left[#3\ \middle|\ #4\right]}{\brackets*{#3}}}}
\NewDocumentCommand{\Varsm}{somo}{\mathrm{Var}\IfValueT{#2}{_{#2}}{} \IfBooleanTF{#1}{#3}{\IfValueTF{#4}{\left[#3\ \middle|\ #4\right]}{\brackets{#3}}}}
\NewDocumentCommand{\Cov}{som}{\mathrm{Cov}\IfValueT{#2}{_{#2}}{} \IfBooleanTF{#1}{#3}{\brackets*{#3}}}
\NewDocumentCommand{\Cor}{som}{\mathrm{Cor}\IfValueT{#2}{_{#2}}{} \IfBooleanTF{#1}{#3}{\brackets*{#3}}}
\NewDocumentCommand{\grad}{e_}{\bm{\nabla}\IfValueT{#1}{_{\!\!#1}\,}}

\NewDocumentCommand{\diag}{som}{\mathrm{diag}\IfValueT{#2}{_{#2}}{}\,#3}

\NewDocumentCommand{\N}{somm}{\mathcal{N}\IfBooleanTF{#1}{\left(}{(}\IfValueT{#2}{#2;}{} #3, #4\IfBooleanTF{#1}{\right)}{)}}
\NewDocumentCommand{\GP}{omm}{\mathcal{GP}(\IfValueT{#1}{#1;}{} #2, #3)}

\newcommand{\va}{\vec{a}}

\newcommand{\vf}{\vec{f}}

\newcommand{\vw}{\vec{w}}

\newcommand{\vs}{\vec{s}}

\newcommand{\veta}{\bm{\eta}}

\newcommand{\vmu}{\bm{\mu}}

\newcommand{\vphi}{\bm{\phi}}
\newcommand{\vpi}{\bm{\pi}}

\newcommand{\vsigma}{\bm{\sigma}}
\newcommand{\vtheta}{\bm{\theta}}

\def\vtau{{\bm{\tau}}}

\newcommand{\mA}{\mat{A}}

\newcommand{\mS}{\mat{S}}

\def\setA{{\mathcal{A}}}

\def\setD{{\mathcal{D}}}

\def\setM{{\mathcal{M}}}
\def\setN{{\mathcal{N}}}

\def\setS{{\mathcal{S}}}

\def\setW{{\mathcal{W}}}

\renewcommand\refname{References}

\makeatletter
\renewcommand{\fnum@figure}{\textbf{Figure \thefigure}}
\renewcommand{\fnum@table}{\textbf{Table \thetable}}
\makeatother

\usepackage{newfloat}
\DeclareFloatingEnvironment[name=Supplementary Figure, fileext=lsf, placement=htbp]{suppfigure}
\usepackage{cleveref}   % load last
\crefname{suppfigure}{Supplementary Fig.}{Supplementary Figs.}
\Crefname{suppfigure}{Supplementary Figure}{Supplementary Figures}

\usepackage{xcolor, soul}
\sethlcolor{red}

\definecolor{grammar_change_color}{HTML}{000000} %black
\definecolor{prev_change_color}{HTML}{000000} %black
\definecolor{review_highlight_color}{HTML}{000000}
\definecolor{format_highlight_color}{HTML}{000000}
\definecolor{newtextcolor}{HTML}{1F77B4} % blue
\definecolor{oldtextcolor}{gray}{0.5}    % gray

\usepackage[normalem]{ulem}
\def\scititle{
	Learning Soft Robotic Dynamics \\with Active Exploration
}
\title{\bfseries \boldmath \scititle}

\author
{Hehui Zheng,$^{1, 3}$ Bhavya Sukhija,$^{2}$ Chenhao Li,$^{2, 3}$ \and
Klemens Iten,$^{2}$ Andreas Krause,$^{2,3}$ Robert K. Katzschmann$^{1,3\ast}$\and
\small$^{1}$Soft Robotics Lab, D-MAVT, ETH Zurich, Zurich, Switzerland\and
\small$^{2}$Learning \& Adaptive Systems Group, D-INFK, ETH Zurich, Zurich, Switzerland\and
\small$^{3}$ETH AI Center, ETH Zurich, Zurich, Switzerland\and
\small$^\ast$Corresponding author. Email: \href{mailto:rkk@ethz.ch}{rkk@ethz.ch}
}
\date{}
\begin{document} 
% \captionsetup[figure]{labelfont={bf},name={Fig.},labelsep=period}
% \captionsetup[table]{labelfont={bf},name={Tab.},labelsep=period}

%highlight grammar fix
\newcommand{\grammar}[1]{\textcolor{grammar_change_color}{#1}}
\newcommand{\review}[1]{\textcolor{prev_change_color}{#1}}
\newcommand{\breview}[1]{\textcolor{review_highlight_color}{#1}}
\newcommand{\format}[1]{\textcolor{format_highlight_color}{#1}}
\newcommand*\diff{\mathop{}\!\mathrm{d}}

%\baselineskip24pt  --------> RONAN: DISABLE DOUBLE LINE SPACING to ease writing <---------

\maketitle 

\begin{abstract}
    Soft robots offer adaptability and safety in unstructured environments, yet their compliant, high-dimensional, and nonlinear dynamics make modeling for control difficult.
    Existing data-driven approaches often fail to generalize, constrained by narrowly focused task demonstrations or inefficient random exploration.
    We introduce an uncertainty-aware active exploration framework that autonomously learns task-agnostic and generalizable dynamics models of soft robotic systems.
    We evaluate this framework on three simulated platforms---a continuum arm, an articulated fish in fluid, and a musculoskeletal leg---and on a real-world pneumatically actuated continuum arm.
    Compared with random exploration and task-specific baseline, our method achieves higher modeling accuracy by actively collecting informative data.
    This improved model quality yields higher returns in zero-shot control on unseen tasks and robust performance under sensing noise, actuation delays, and material nonlinearities.
    These results demonstrate that uncertainty-driven active exploration enables data-efficient dynamics learning across soft robotic morphologies, supporting autonomous and adaptable control for compliant robots.
\end{abstract}
% \textbf{Short title:} Exploring Soft Dynamics\\\\
% \textbf{Summary:} Active uncertainty-driven exploration lets soft robots autonomously learn accurate, reusable dynamics for diverse tasks.

\newpage

\section*{\format{Introduction}}
    Robotics has achieved progress in precision, speed, and reliability, driven by advances in rigid mechanisms. 
    However, as robots expand into homes and shared workspaces with humans, rigid structures pose inherent challenges for safe physical interaction~\cite{lasota2017survey}. 
    This growing need for human-centered robotics has fueled interest in soft robots, whose deformable and compliant bodies enable safe contact, adaptive grasping, and robust performance in unstructured environments~\cite{yasa2023overview}.
    % Soft robots contain highly deformable and compliant structures that offer compelling advantages in adaptability, dynamic manipulation, and safe human interaction~\cite{yasa2023overview}.
    Yet these same characteristics make their behavior challenging to model and control~\cite{della2023model}.
    Unlike rigid-body systems described by low-dimensional ordinary differential equations, soft robots exhibit complex, high-dimensional dynamics driven by continuous deformation, distributed compliance, and nonlinear material properties~\cite{hawkes2017soft,du2021underwater, wang2022control, yang2024nonlinear, liu2025exploring}.
    They lack well-defined coordinates or rigid links, rendering classical kinematic and dynamic formulations imprecise~\cite{spinelli2022unified, kazemipour2022adaptive, fischer2023dynamic}.
    Instead, their motion and interaction behavior emerges from nonlinear partial differential equations and rich physical phenomena, such as viscoelasticity, hysteresis, and nonlocal actuation responses~\cite{qu2023modeling, qin2024modeling}, making accurate modeling difficult to achieve using analytical tools alone.

    Physics-based modeling has long provided foundational insights into soft robotic behavior, but existing approaches face significant limitations.
    Analytical models such as piecewise constant curvature (PCC) offer interpretability and tractability, yet often fail to capture elongation, anisotropy, and non-constant curvature observed in real hardware, resulting in non-negligible tip position errors~\cite{toshimitsu2021sopra, zheng2024vision}.
    % Increasing the number of PCC segments or supporting variable-length sections improves accuracy but quickly increases computational cost.
    Finite element methods (FEM) captures nonlinear material behavior and distributed deformation with high fidelity, but tendon-driven or fiber-reinforced structures require complex parameterization and dense meshing that sharply increase computational burden~\cite{kokubu2024development}.
    Simplifying internal reinforcements using surrogate structures such as stiff spring rings~\cite{cangan2022model} can reduce complexity, but often impair dynamic realism~\cite{chen2025accelerated}.
    As a result, FEM-based models frequently perform well in quasi-static regimes but struggle under fast actuation, high load, or complex contact conditions.
    Reduced-order models aim to balance fidelity and tractability~\cite{katzschmann2019dynamically, tonkens2021soft, stella2023piecewise}, yet typically sacrifice generality and robustness across environments.

    Data-driven approaches have therefore gained momentum as an alternative for capturing soft robot dynamics.
    Neural networks, Gaussian processes, and Koopman operator-based models have demonstrated strong performance in tasks ranging from locomotion to shape estimation~\cite{gillespie2018learning, thuruthel2019soft, sedal2021comparison, chen2024data}.
    However, their generalizability is fundamentally constrained by the quality and diversity of the training data~\cite{raissi2019physics,li2025robotic,laschi2023learning}.
    Task-specific demonstrations produce narrow, overfitted models that struggle outside the training distribution, while random exploration fails to efficiently cover the vast and sparsely reachable state--action spaces characteristic of soft systems.
    Hybrid approaches that augment analytical models with learned residuals~\cite{choi2023learning, levy2024learning} or incorporate physical priors into Bayesian deep learning~\cite{rothfuss2024bridging} improve sample efficiency and interpretability, but remain bottlenecked by data coverage.
    Simulation-based augmentation is further limited by discrepancies between simulated and physical materials, unmodeled interactions, and slow computational throughput~\cite{gao2024sim}, reducing the reliability of synthetic data for real-world generalization.

    To overcome these limitations, research in robot learning has increasingly emphasized active exploration as a way to autonomously acquire informative data.
    Approaches based on intrinsic motivation, information gain, or uncertainty sampling have shown promise in collecting diverse data for reinforcement learning, system identification, and model-based control~\cite{aubret2019survey, sekar2020planning, latyshev2023intrinsic, kim2023bridging}.
    These methods improve generalization in rigid-body systems and in simulated environments~\cite{sukhija2024optimistic}.
    Curiosity-driven and ensemble-based uncertainty estimation techniques have also been used in high-dimensional or underactuated robots~\cite{pathak2017curiosity, bechtle2020curious, seyde2020learning, fasel2022ensemble, kim2023bridging, li2025offline}.
    However, relatively little work has explored such strategies in soft robotics, where high-dimensional deformation, sparse sensing, compliant materials, and coupled dynamics create exploration challenges that differ substantially from those in rigid systems~\cite{liu2023learning, jitosho2023reinforcement, shao2025selfattention}.
    Existing studies remain largely simulation-only, leaving open questions about robustness under real-world noise, material variability, and actuation uncertainty.
    To date, no systematic investigation has examined uncertainty-driven exploration for physically compliant soft robotic platforms.

\begin{figure}[!t]
        \centering
        \includegraphics[width=5.3in]{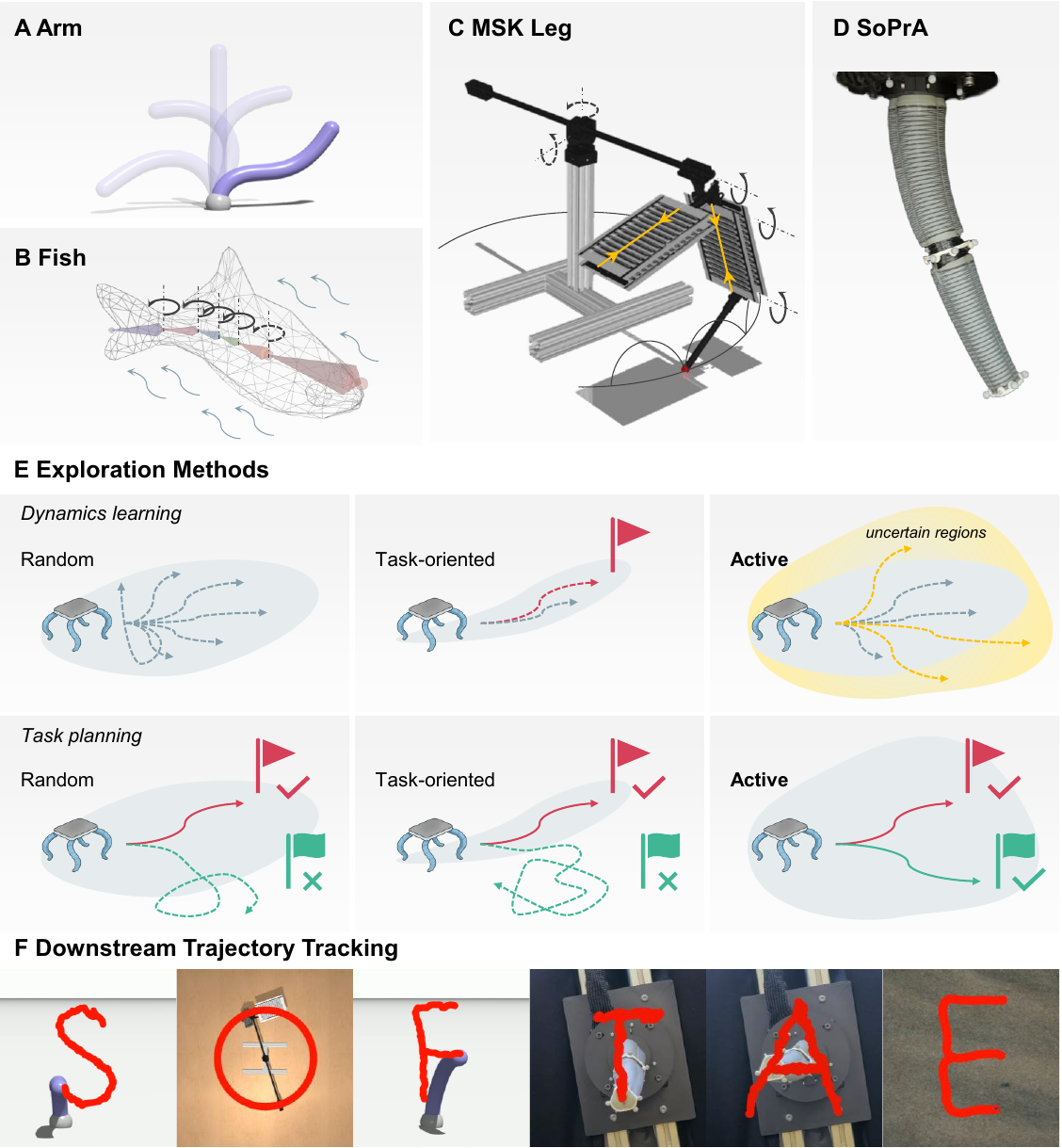}
        \caption{\textbf{Learning soft robotic dynamics with active exploration.}   
        (\textbf{A}--\textbf{D}) Soft robotic platforms studied in this work: (A) soft continuum arm, (B) deformable fish in fluid, (C) musculoskeletal (MSK) leg with electrohydraulic actuation (yellow arrows), and (D) real-world pneumatically actuated arm (SoPrA).
        Their nonlinear, high-dimensional, and deformable dynamics pose significant challenges for generalizable model learning.
        (\textbf{E})~We address this with \textsc{SoftAE}, an active exploration strategy that autonomously collects informative and diverse data for training dynamics models.
        During dynamics learning (top), \textsc{Random} exploration is unguided, and task-oriented method overfits.
        \textsc{SoftAE} instead targets regions of high model uncertainty, improving data coverage and model accuracy.
        This enables robust generalization to diverse downstream tasks (bottom), as (\textbf{F}) demonstrated across all platforms.}
        \label{fig:1_overview}
    \end{figure}

    These gaps motivate our focus on autonomously learning task-agnostic and general-purpose dynamics models for soft robots. 
    We introduce \textsc{SoftAE} \review{(\textit{Soft-robot Active Exploration})}, an uncertainty-driven active exploration framework for learning soft robotic dynamics (\format{Fig.}~\ref{fig:1_overview} \format{and Supplementary Movie~1}).
    Rather than relying on task demonstrations or random perturbations, \textsc{SoftAE} leverages model uncertainty as an intrinsic motivation signal to guide data collection toward regions of the state--action space where the current dynamics model is least certain. 
    This targeted strategy enables efficient and comprehensive coverage of the robot’s reachable workspace.
    The resulting dynamics models are task-agnostic and highly generalizable, supporting zero-shot planning and control for previously unseen tasks without retraining, which is an essential property given the cost and effort required to obtain additional real-world data for soft robots.

    The primary contributions of this work are as follows.
    (1) We propose \textsc{SoftAE}, an uncertainty-guided active exploration framework that autonomously targets informative trajectories to achieve efficient coverage of system dynamics. \review{Beyond applying ensemble uncertainty to soft robotics, \textsc{SoftAE} introduces an optimistic exploration mechanism that plans over plausible dynamics within the learned uncertainty set, rather than merely propagating or sampling model uncertainty. This converts epistemic uncertainty into directed exploration of informative, hard-to-reach deformation regimes that are difficult to discover through random actuation or task-specific optimization.}
    (2) We demonstrate that the models learned through \textsc{SoftAE} are task-agnostic, enabling zero-shot planning on downstream tasks without task-specific training or fine-tuning.
    (3) We provide empirical validation both in simulation and on real soft robotic hardware, showing that across diverse systems \textsc{SoftAE} improves data coverage, dynamics model accuracy, and zero-shot control performance compared with random exploration and task-specific model-based baseline. 
    These experiments highlight not only the feasibility of our method but also its robustness to real-world complexities such as sensor noise, actuation uncertainty, and material nonlinearities. 
    Overall, our work provides a practical and scalable framework for building reusable soft robot models \breview{with task-agnostic active exploration}.    

\section*{Results}

    \subsection*{\breview{Soft Robotic Platforms}}

    We evaluate the proposed active exploration pipeline across both simulated and real-world soft robotic systems with highly nonlinear dynamics (\format{Fig.}~\ref{fig:1_overview}, A--D\format{, Supplementary Movies~2--5}). 
    Specifically, we assess whether the learned models (i) capture the full behavioral range of each system, (ii) enable accurate zero-shot planning across diverse downstream tasks, and (iii) can be obtained reliably through a unified pipeline across diverse soft robotic morphologies and control regimes.
    We first demonstrate results in simulation, and then validate on real hardware.

    The simulated experiments span three soft robotic systems (\format{Fig.}~\ref{fig:1_overview}, A--C): a continuum arm modeled as a Cosserat rod, an articulated fish with deformable skin swimming in water, and a musculoskeletal leg actuated by electrohydraulic muscles and a direct current (DC) motor. 
    These environments encompass distinct dynamic behaviors---continuous elastic deformation, two-way fluid--structure interaction, and compliant muscle-actuated motion with ground contact---and demand fundamentally different actuation modalities.
    Their associated tasks also vary substantially, from fixed-base reaching to aquatic locomotion and contact-rich hopping.
    To examine real-world applicability, we apply the same exploration strategy to the pneumatically actuated SoPrA arm~\cite{toshimitsu2021sopra} (\format{Fig.}~\ref{fig:1_overview}D). 
    Across all systems, the learned dynamics models are evaluated on multiple downstream tasks without task-specific retraining.

    \subsection*{Baseline Exploration and Learning Methods}
    We compare our active exploration method (\textsc{SoftAE}) against two representative baselines (\format{Fig.}~\ref{fig:1_overview}E). 
    As an undirected baseline, \textsc{Random} samples actions uniformly from the action space, providing a reference for exploration without uncertainty guidance. 
    \review{We emphasize that uniform sampling in the action space does not translate into uniform coverage of the state--action space: the states actually visited are an emergent consequence of the rolled-out actions and the system dynamics, and for compliant, underactuated systems they typically concentrate in a small region of the reachable space rather than spanning it~\cite{sekar2020planning, sukhija2024optimistic, buisson2020actively}.}
    To assess the value of task-agnostic exploration for downstream generalization, we also compare against \textsc{H-UCRL}~\cite{curi2020efficient}, a model-based reinforcement learning (RL) method trained specifically on individual downstream tasks. 
    All methods employ probabilistic ensembles (PEs)~\cite{lakshminarayanan2017simple} to model dynamics and use model predictive control (MPC) with the improved Cross-Entropy Method (iCEM) optimizer~\cite{pinneri2021sample} for action planning.

    \subsection*{Zero-shot Task Performance Across Simulated Soft Robotic Systems}
        To systematically evaluate generalization, we consider three simulated systems of diverse physical and control complexity: a continuum arm, an articulated fish with deformable skin immersed in fluid, and a musculoskeletal leg actuated by electrohydraulic muscles and a DC motor (\format{Fig.}~\ref{fig:1_overview}, A--C).
        \cref{tab:1_sim_env_summary} summarizes the task definitions and state--action space dimensions, with full \review{environment} descriptions provided in Methods.
        \review{Per-environment state--action decompositions and reward formulations are provided in Supplementary} \format{Tables~\ref{tabS:1_arm_state_action}--\ref{tabS:1_rewards}}.
        
        \begin{table}[t]
            \centering
            \caption{\textbf{Simulated soft robotic environments for dynamics learning.} 
            The table summarizes three environments investigated in this work: soft continuum arm, articulated fish with deformable skin in fluid, and musculoskeletal leg, together with their associated tasks. 
            For each environment, the state space dimensions correspond to the size of the observation vector, while the action space dimensions denote the number of independent actuation signals. 
            Detailed task rewards and decomposition of state--action space are provided in Supplementary \format{Tables~\ref{tabS:1_arm_state_action}--\ref{tabS:1_rewards}}.
            % \cite{methods}.
            % These environments differ in actuation type, state and action dimensionality, physical complexity, and task objectives.
            }
            \label{tab:1_sim_env_summary}
            \begin{tabular}{@{}llcc@{}}
                \toprule
                \textbf{Environment}        
                                            % & \textbf{Simulator}
                                            & \textbf{Task} 
                                            & \begin{tabular}[c]{@{}l@{}}\textbf{State Space}\\ \textbf{Dimensions}\end{tabular}
                                            & \begin{tabular}[c]{@{}l@{}}\textbf{Action Space}\\ \textbf{Dimensions}\end{tabular} \\ \midrule
                \multirow{2}{*}{Soft Continuum Arm}        
                                            % & \multirow{2}{*}{Elastica} 
                                            & \textit{(i)} reach close target                 
                                            & $58$ & $12$ \\ 
                                            & \textit{(ii)} reach far target 
                                            & $58$ & $12$\\
                \midrule
                \multirow{4}{*}{Articulated Fish in Fluid}       
                                            % & \multirow{4}{*}{FishGym}
                                            & \textit{(iii)} swim along +x direction   
                                            & $15$ & $4$  \\
                                            & \textit{(iv)} swim along -x direction   
                                            & $15$ & $4$  \\
                                            & \textit{(v)} swim to target                      
                                            & $18$ & $4$ \\
                                            &\textit{(vi)} swim away from target              
                                            & $18$ & $4$ \\ 
                \midrule
                \multirow{2}{*}{Musculoskeletal Leg}    
                                            % & \multirow{2}{*}{MuJoCo} 
                                            & \textit{(vii)} hop counterclockwise  
                                            & $10$ & $5$                    \\
                                            & \textit{(viii)} hop clockwise
                                            & $10$ & $5$         \\ 
                \bottomrule
                \end{tabular}%
        \end{table}

        We evaluate downstream performance across eight tasks using dynamics models learned from exploration alone, without any task-specific fine-tuning.
        Our active exploration method (\textsc{SoftAE}) and the \textsc{Random} baseline are tested in a zero-shot setting, while the task-specific model-based RL baseline (\textsc{H-UCRL}) is trained only on tasks \textit{(i)}, \textit{(iii)}, \textit{(v)}, and \textit{(vii)}; the remaining tasks serve as held-out generalization tests. 
        Each method is evaluated over 10 random seeds for tasks \textit{(i)}--\textit{(ii)} and 5 seeds for the others.

        \begin{figure}[!t]
            \centering
            \includegraphics[width=6.2in]{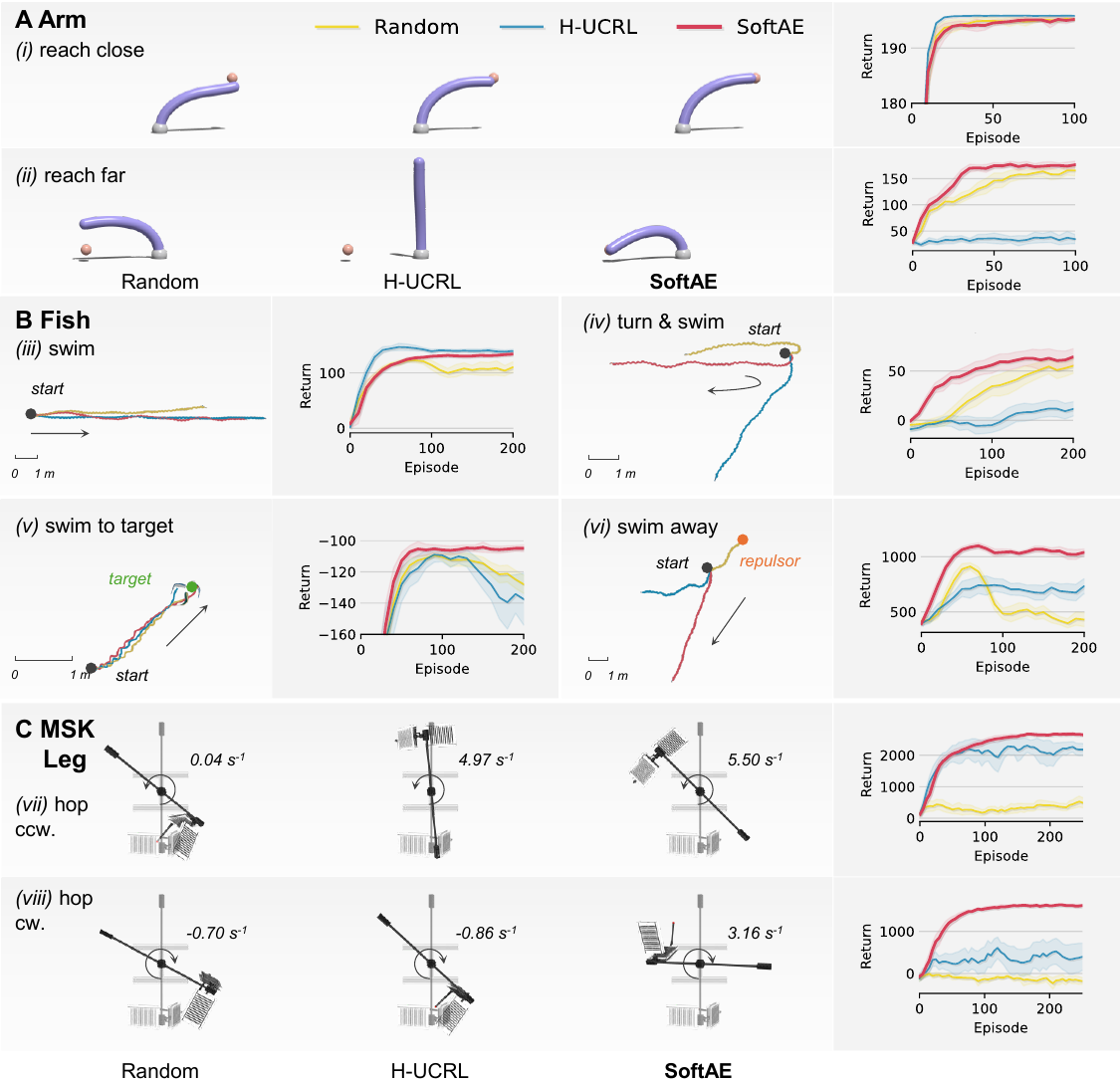}
            \caption{\textbf{Active exploration enables robust zero-shot task performance across diverse soft robotic systems.}
            We evaluate \textsc{SoftAE} against baseline exploration strategies across three soft robotic platforms:
            (\textbf{A}) a soft continuum arm,
            (\textbf{B}) a deformable fish swimming in fluid, and
            (\textbf{C}) a musculoskeletal (MSK) leg actuated by electrohydraulic muscles and a DC motor.
            \textsc{H-UCRL}, a task-specific model-based baseline, is trained only on tasks \textit{(i)}, \textit{(iii)}, \textit{(v)}, and \textit{(vii)}, while the remaining tasks are held out as unseen.
            Across tasks, \textsc{SoftAE} consistently produces task-aligned and physically plausible behaviors, while \textsc{Random} and \textsc{H-UCRL} often fail to generalize to the hard or unseen scenarios.
            Returns are averaged over 10 random seeds for tasks \textit{(i)} and \textit{(ii)}, and 5 seeds for the remaining tasks\breview{; shaded regions denote $\pm$ one standard error of the mean}.
            }
            \label{fig:2_sim_tasks}
            \vspace{-1em}
        \end{figure}

        As shown in \format{Fig.}~\ref{fig:2_sim_tasks}, \textsc{SoftAE} substantially outperforms \textsc{Random}, particularly in environments where effective behavior depends on coordinated actuation under nonlinear or delayed dynamics. 
        For instance, swimming requires synchronized spine undulation against delayed hydrodynamic feedback, while hopping demands precise muscle-motor coordination and accurate timing of ground contact. 
        Lacking such structured patterns, \textsc{Random} exploration seldom produces meaningful motion, preventing it from discovering viable behaviors and leading to failure on tasks \textit{(v)}--\textit{(viii)}. 

        Across all tasks, \textsc{SoftAE} matches or exceeds the performance of \textsc{H-UCRL} on tasks it was trained \breview{on}, and outperforms it on held-out tasks where \textsc{H-UCRL} fails to generalize. 
        \textsc{SoftAE} \breview{also} surpasses \textsc{H-UCRL} even on tasks \textit{(v)} and \textit{(vii)} (fish swimming and leg hopping), despite \textsc{H-UCRL} being trained directly on these objectives. 
        These results underscore the limitations of task-specific exploration, which may overlook coordinated control patterns not easily discovered by local optimization around a single objective. 
        In contrast, \textsc{SoftAE}'s uncertainty-driven exploration strategy collects diverse data that enables robust, zero-shot control across the three distinct soft robotic platforms and tasks evaluated here. 
    
    \subsection*{Data Diversity Through Active Exploration}
        
\begin{figure}[]
            \centering
            \includegraphics[width=0.9\textwidth]{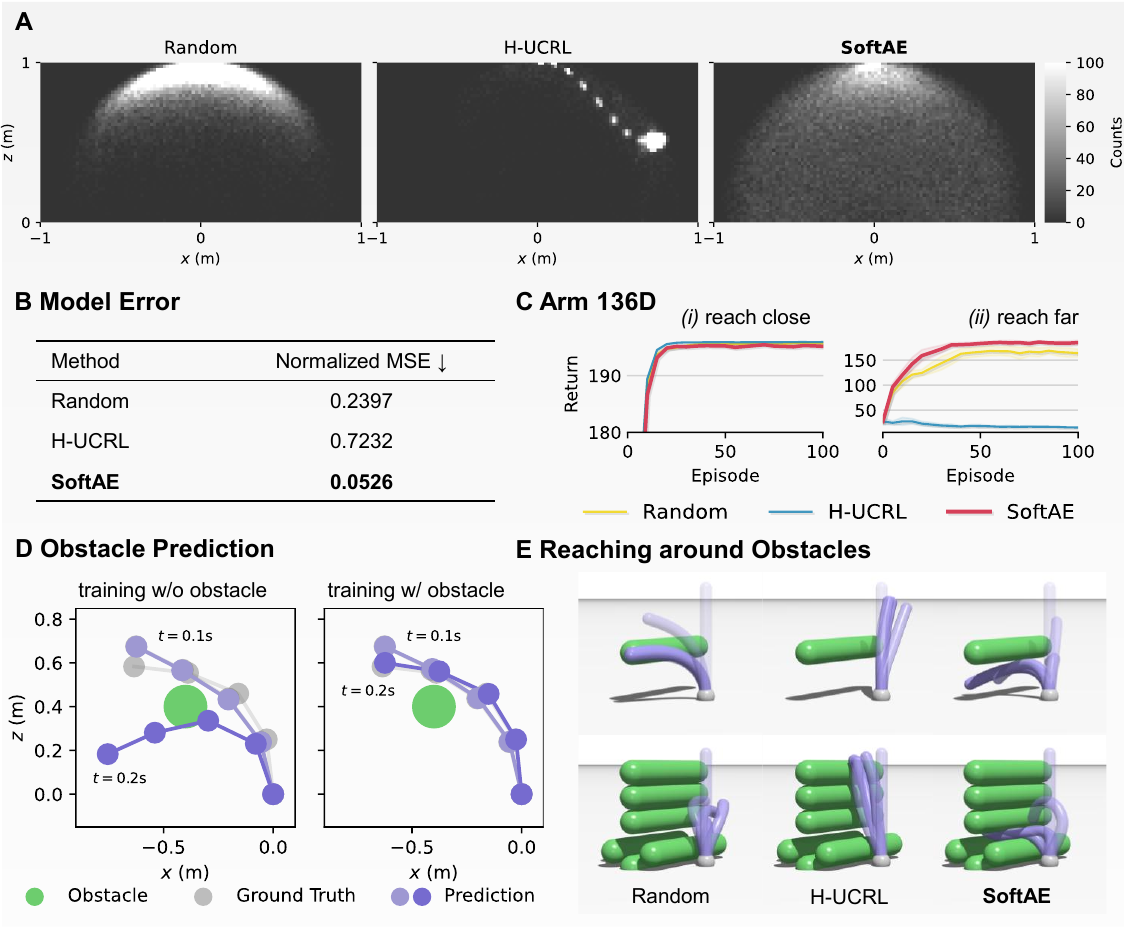}
            \caption{
            \textbf{Active exploration improves data coverage, model accuracy, scales to high-dimensional dynamics, \review{and captures soft--rigid contact.}}  
            \textbf{(A)} Spatial coverage of collected data for the soft continuum arm, with tip positions projected onto the $x$--$z$ plane. Heatmaps show visitation frequency per spatial bin. \textsc{SoftAE} explores more broadly and uniformly than \textsc{Random} and \textsc{H-UCRL}.
            \textbf{(B)} Normalized mean squared error (MSE) of learned dynamics models on a held-out set of 17,451 transitions. \textsc{SoftAE} achieves the lowest error.
            \textbf{(C)} Return curves for reaching \textit{(i)} close and \textit{(ii)} far targets on a soft arm with a 136-dimensional state space. \textsc{SoftAE} maintains high performance in this high-dimensional setting, matching \textsc{H-UCRL} on its trained task and outperforming baselines on the unseen task. Returns are averaged over 10 seeds\breview{; shaded regions denote $\pm$ one standard error of the mean}. 
            \review{(\textbf{D}) Autoregressive 10-step prediction (from $t = 0.1$\,s to $t = 0.2$\,s) of learned dynamics models in an environment with a rigid spherical obstacle (green). A model trained without obstacle exposure (left) predicts the arm passing through the obstacle, while a model trained in the obstacle environment (right) accurately reproduces the contact-induced deformation.} 
            \review{(\textbf{E}) Downstream reaching tasks with rigid obstacles, where \textsc{SoftAE} and \textsc{Random} explore in the obstacle environment and \textsc{H-UCRL} is trained on task \textit{(i)}. Top: reaching under a horizontal bar. Bottom: reaching the opposite side of a vertical wall through a gap near its base. Only \textsc{SoftAE} consistently succeeds at both contact-rich tasks.}
            }
            \label{fig:3_model_eval}
            \vspace{-1em}
        \end{figure}

        The strong downstream performance of \textsc{SoftAE} suggests that its advantage arises from collecting training data with broader and more informative coverage than baselines. 
        To test this hypothesis, we examine the exploration behavior in the continuum arm environment (\format{Fig.}~\ref{fig:1_overview}A). 
        Because the full 58-dimensional state space cannot be visualized directly, we project the arm's 3D tip position onto a 2D plane and aggregate visiting frequencies into workspace heatmaps (\format{Fig.}~\ref{fig:3_model_eval}A).

        \review{Both baselines fail to cover the workspace, but for different reasons.}
        Task-specific \textsc{H-UCRL} gathers data narrowly along its training trajectory, underscoring its lack of task-agnostic exploration.
        \textsc{Random} \review{instead} concentrates samples near the rest state\review{, a direct consequence of the arm's compliance.
        Because \textsc{Random} resamples actions independently at each control step, the applied torques tend to cancel rather than accumulate, and the arm relaxes toward rest before it can travel far.
        Configurations involving large deformation are thus reachable only through temporally correlated, multi-step actuation sequences, which independent uniform action sampling produces with vanishing probability.} 
        \textsc{SoftAE}\review{, by contrast,} achieves far more uniform coverage across the reachable workspace.
        
        \review{Because \textsc{Random}'s failure is a structural limitation of undirected sampling rather than a matter of sampling efficiency, it cannot be overcome simply by collecting more data.
        As shown in Supplementary \format{Fig.~\ref{figS:1_random_extended}}, even with five times the exploration budget, \textsc{Random} only approaches the other methods on the easier task \textit{(i)} and still plateaus well below \textsc{SoftAE} on the harder task \textit{(ii)}.}
        \review{Together,} these patterns show that active exploration systematically drives the robot toward underexplored regions, producing broader and more balanced data distributions. 
        Such diversity in turn provides richer supervision for dynamics learning, which we next assess through model accuracy.
        
    \subsection*{Model Accuracy with Collected Data}
        To quantify how exploration diversity impacts dynamics model learning, we evaluate the accuracy of dynamics models trained on data from each method.
        Specifically, we compute the mean squared error (MSE) between predicted and ground-truth next states on a held-out validation set consisting of rollouts toward 500 uniformly sampled target positions across the workspace.
        Rollouts are truncated once the target is reached to avoid oversampling near-goal regions, resulting in a total of 17,451 transitions. 
        All state dimensions are normalized by their standard deviation before computing the MSE.

        As reported in \format{Fig.}~\ref{fig:3_model_eval}B, \textsc{SoftAE}'s broader and more balanced exploration \breview{reduces the prediction error by about 4.6-fold relative to \textsc{Random} and 14-fold relative to \textsc{H-UCRL}} across the reachable workspace. 
        In other words, by targeting regions of high uncertainty, active exploration improves data coverage, \breview{which} translates into more accurate dynamics models \breview{and, in turn,} the \breview{higher} zero-shot task \breview{returns}.

    \subsection*{Scalability to High-Dimensional Dynamics}

        For soft robotic systems with continuous and complex deformation, the resulting state representations are often high-dimensional, posing challenges for model learning and planning.
        To assess whether our exploration strategy scales to such settings, we revisit the continuum arm from tasks \textit{(i)} and \textit{(ii)} and increase the observation resolution. 
        Instead of tracking five discrete points along the arm, we extract features from eleven evenly spaced points, resulting in a 136-dimensional state space while keeping the action space and task objectives unchanged.

        As shown in \format{Fig.}~\ref{fig:3_model_eval}C, \textsc{SoftAE} maintains high performance on both reaching tasks, achieving fast learning and high final reward despite the increased dimensionality. 
        In contrast, \textsc{Random} baseline performance slightly degrades, highlighting the value of uncertainty-aware exploration in guiding data collection in large state spaces. 

    \subsection*{\review{Generalization to Soft--Rigid Contact Interactions}}

        \review{The simulated environments evaluated thus far include interactions between the soft body and its surroundings: the articulated fish operates within a two-way coupled fluid--structure simulation, and the musculoskeletal leg involves repeated ground contact at the foot. 
        These results \breview{indicate} that \textsc{SoftAE} can learn dynamics shaped by external coupling. 
        However, soft--rigid contact between a deformable body and a rigid environmental obstacle poses a further challenge.
        Contact can occur at any point along the arm depending on its configuration, and the resulting deformation is shaped jointly by the arm's compliance and the obstacle geometry. 
        To examine this regime, we revisit the simulated soft continuum arm and place rigid obstacles within its reachable workspace, so that the arm must interact with the objects during both exploration and downstream control.}
        
        \review{We first ask whether the learned dynamics model can capture soft--rigid contact when contact transitions are present in the training data. 
        We compare two models trained with \textsc{SoftAE}: one in an obstacle-free environment, the other with a rigid spherical obstacle. 
        Starting from the state immediately preceding contact ($t = 0.1$\,s), we perform an autoregressive rollout (10 steps at 100\,Hz) and compare the predicted arm configuration at $t = 0.2$\,s against ground truth (\format{Fig.}~\ref{fig:3_model_eval}D). 
        The obstacle-free model has no notion of the contact and predicts the arm passing through the obstacle, whereas the model trained with the obstacle accurately reproduces the contact-induced deformation. 
        This confirms that neural-network dynamics models can represent the discontinuous behavior arising from soft--rigid interaction, provided such transitions are present in the training data.}

        \review{Next, we assess whether exploration can reliably collect data covering both the contact regions and the workspace occluded by an obstacle.
        We evaluate downstream reaching tasks in which the target is occluded by a rigid obstacle: a horizontal bar to reach beneath, and a vertical wall with a gap near its base, through which the arm must pass to reach the opposite side (\format{Fig.}~\ref{fig:3_model_eval}E). 
        \textsc{SoftAE} and \textsc{Random} explore in the obstacle environment, while \textsc{H-UCRL} is trained there directly on task \textit{(i)}, which does not require interaction with the obstacle. 
        As shown in \format{Fig.}~\ref{fig:3_model_eval}E, only \textsc{SoftAE} consistently succeeds at both contact-rich tasks. 
        \textsc{Random} does collect contact data on the exposed surfaces of the obstacles, but fails to discover the coordinated deformations required to maneuver around or through them. 
        \textsc{H-UCRL} does not generalize to obstacle-aware reaching. 
        These results indicate that the benefits of uncertainty-driven exploration extend beyond free-space dynamics to soft--rigid contact, broadening the practical scope of the framework.}
    
    \subsection*{Real-World Active Exploration with a Pneumatically Actuated Arm}

        \begin{figure}[!ht]
            \centering
            \includegraphics[width=5.5in]{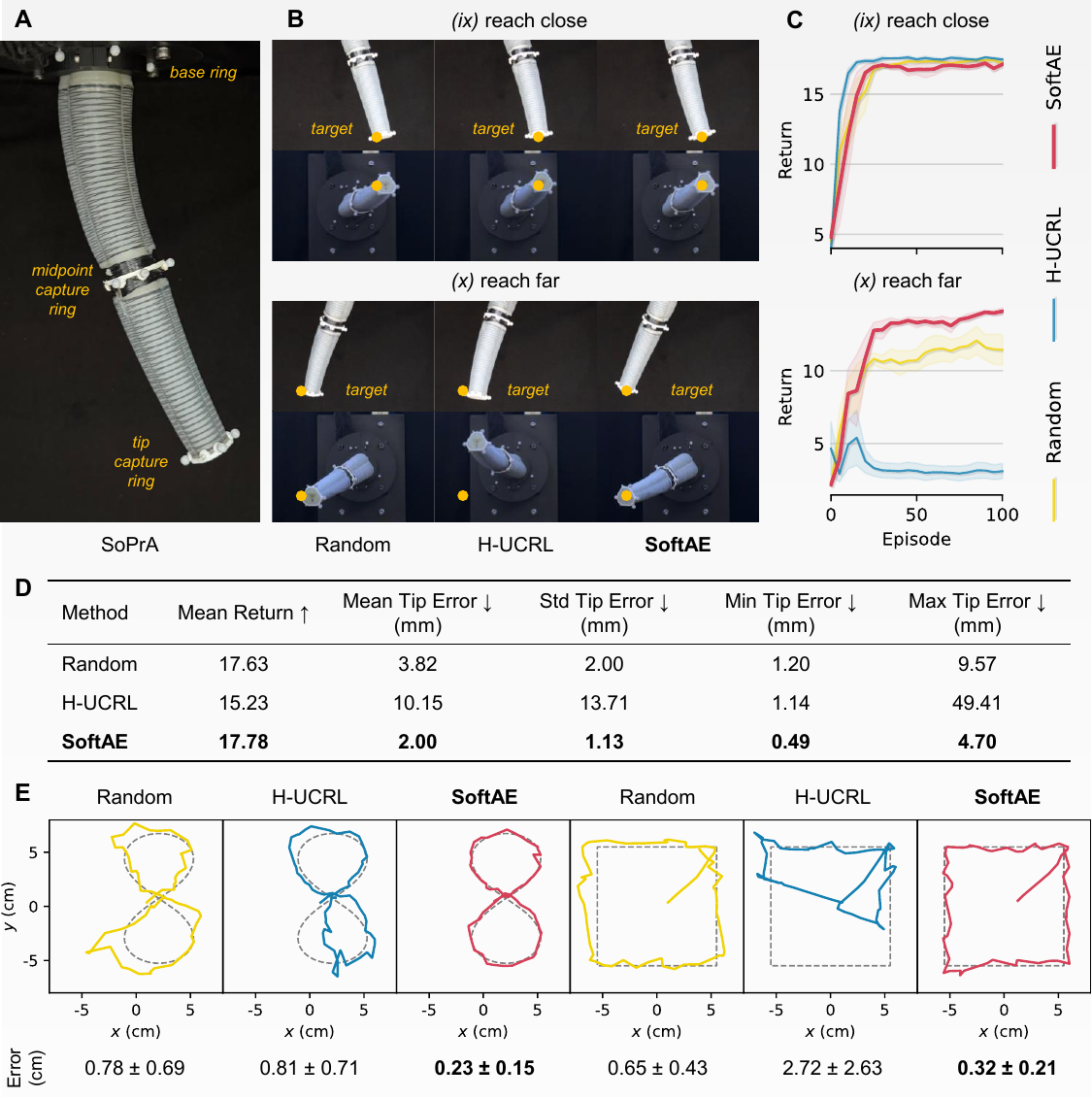}      
            \caption{\textbf{Real-world dynamics learning and task performance on a pneumatically actuated soft arm.}
                (\textbf{A})~SoPrA: a fiber-reinforced, pneumatically actuated continuum arm with motion capture via marker rings.
                (\textbf{B})~Reaching tasks for \textit{(ix)} close and \textit{(x)} far targets. Top row: front view; bottom row: bottom view. Models are trained using \textsc{Random}, \textsc{H-UCRL} (task-specific, trained only on \textit{(ix)}), or \textsc{SoftAE} (ours). Yellow dots indicate target positions.
                (\textbf{C})~Return curves show all methods perform similarly on task \textit{(ix)}, while \textsc{SoftAE} outperforms on the harder, unseen task \textit{(x)}. Returns are averaged over 3 random seeds, shaded regions denote standard error of the mean.
                (\textbf{D})~Quantitative evaluation over the same set of 20 random targets shows that \textsc{SoftAE} achieves the highest mean return and lowest tip position error compared to baselines.
                (\textbf{E})~Trajectory tracking performance on two reference trajectories: square (top row) and lemniscate (bottom row). \textsc{SoftAE} tracks both shapes more accurately than baselines.
                }
            \label{fig:4_real_sopra}
        \end{figure}

        Having validated \textsc{SoftAE} across multiple simulated systems, we next assess its performance on a physical soft robotic platform.
        Modeling such continuum systems in hardware is particularly challenging due to continuous deformation, nonlinear material properties, and coupled pneumatic actuation.
        To examine whether our exploration framework can handle these complexities, we apply it to the pneumatically actuated SoPrA arm~\cite{toshimitsu2021sopra}, a two-segment continuum arm constructed from a continuous silicone shell with six internal chambers \breview{(Supplementary Movie~5)}.
        Each segment is assembled by combining three individually fabricated fiber-reinforced chambers, and all six chambers can be independently pressurized to generate complex, spatially distributed deformations for 3D motion.
        This design provides a representative example of high-compliance soft manipulators commonly used in human-robot interaction, soft grasping, and wearable robotics~\cite{abozaid2024soft, paterno2023soft}. 
        
        The arm is actuated through differential pressure commands, forming a 6-dimensional action space corresponding to the pressure change in each chamber.
        Its state is captured using motion-capture markers on rings mounted at the base, midpoint, and tip (\format{Fig.}~\ref{fig:4_real_sopra}A).
        For the midpoint and tip rings, we observe the 3D position, linear velocity, orientation (as a quaternion), and angular velocity relative to the base ring.
        Together with the absolute pressure of all six chambers, this yields a 32-dimensional state space \review{(full decomposition in Supplementary \format{\cref{tabS:1_sopra_state_action})}}.

        \review{We adopt a hierarchical control architecture: the high-level control planner operates at 1\,Hz for both exploration and downstream task control, providing pressure setpoints to a low-level pressure controller that regulates the six chambers at 10\,Hz. This separation reflects the intrinsic bandwidth of pneumatic soft actuation, where pressure dynamics evolve on the order of hundreds of milliseconds due to compressible-fluid response and viscoelastic behavior of the silicone shell. Planning at 1\,Hz matches this physical timescale, while the low-level controller ensures that commanded pressures are tracked smoothly between planning steps.}

        For downstream evaluation, we define two 3D target-reaching tasks, mirroring the simulated continuum arm tasks: task \textit{(ix)} reaches a nearby target, while task \textit{(x)} reaches near the workspace boundary, requiring greater deformation. 
        Robust zero-shot planning in this real-world setting must contend with sensor noise, actuation delays, and physical nonidealities such as compliance, hysteresis, and pneumatic variability.

        Due to hardware constraints, we evaluate real-world performance using 3 random seeds. 
        After a fixed exploration budget, \textsc{SoftAE} and \textsc{Random} each learn a dynamics model, and \textsc{H-UCRL} is trained directly on task \textit{(ix)}.
        As shown in \format{Fig.}~\ref{fig:4_real_sopra}, B and C, all methods achieve comparable performance on the easier task \textit{(ix)}, where even random exploration provides adequate data coverage near the rest configuration. 
        However, on the more challenging task \textit{(x)}, which involves reaching toward the edge of the workspace, \textsc{SoftAE} outperforms both \textsc{Random} and \textsc{H-UCRL}. 

        \review{Beyond these model-based comparisons, we also evaluate an online model-free reinforcement learning baseline, Soft Actor-Critic (SAC,~\cite{haarnoja2018sac}), trained directly on real-world interaction without a learned dynamics model. 
        As reported in Supplementary \format{Methods} (\format{Supplementary Fig.}~\ref{figS:5_sac}), SAC \breview{trained on goal-conditioned reaching} fails to make meaningful progress within twice the interaction budget granted to the model-based methods. 
        \breview{This underscores the data efficiency of model-based exploration at budgets feasible on physical soft robots.}}

        To \review{further} assess generalization, we evaluate \review{the three model-based} methods on the same set of 20 randomly sampled 3D target positions within the arm’s reachable workspace (\format{Fig.}~\ref{fig:4_real_sopra}D). 
        \textsc{SoftAE} achieves the highest mean return and the lowest average tip error of \SI{2}{mm}, indicating both time-efficient reaching behavior and high positional accuracy across diverse target positions.
        This level of error is well within practical tolerances for soft arm manipulation tasks such as reaching and positioning, where centimeter-level accuracy is usually sufficient due to the compliance and safety of the manipulator. 
        While \textsc{Random} exploration performs reasonably well on close targets, it suffers from higher variability due to inconsistent coverage of the workspace. 
        \textsc{H-UCRL} exhibits the highest maximum tip error (\SI{49.41}{mm}), over ten times that of \textsc{SoftAE} (\SI{4.70}{mm}), underscoring the brittleness of task-specific training when deployed beyond its scope.
        In contrast, \textsc{SoftAE}’s uncertainty-guided exploration yields a more robust and generalizable model that maintains accuracy throughout the workspace. 
        Such consistency is critical for real-world deployment of soft robotic systems, where new tasks may arise without retraining and safety margins depend on reliable worst-case performance.
        Finally, we evaluate performance on continuous trajectory tracking, using both square and lemniscate reference paths (\format{Fig.}~\ref{fig:4_real_sopra}E).
        \textsc{SoftAE} tracks both trajectories more accurately than the baselines, which often fail in regions where their models lack sufficient training coverage.

\section*{Discussion}
    % \subsection*{Key Findings and Implications}
    This work addresses the core challenge of learning generalizable dynamics models for soft robotic systems, whose behaviors are inherently high-dimensional, nonlinear, and history-dependent. 
    We present an active exploration framework that leverages epistemic uncertainty to guide data collection toward informative, underexplored regions of the state--action space. 
    Through systematic evaluation across three simulated platforms, including a soft continuum arm, an articulated fish robot, and a musculoskeletal leg, we demonstrate that this approach enables efficient and scalable model learning across fundamentally different robotic regimes.
    The resulting models support robust zero-shot control across multiple downstream tasks without task-specific retraining. 
    We further validate the method on a pneumatically actuated continuum arm in the real world, where it maintains reliable performance under sensing noise, actuation delays, and complex material behaviors.

    These results highlight the importance of data-efficient exploration for modeling and control of soft robots, where accurate simulation is often unavailable, a large amount of data is required to capture the continuous high-dimensional state--action space, and real-world data collection is expensive and time-consuming. 
    Unlike random exploration, which fails to scale with system complexity, or task-specific model-based approaches that overfit to narrow task regions, our framework autonomously directs data collection towards underexplored regions, enabling broader model coverage with fewer samples. 
    Moreover, because soft robots often exhibit continuous deformation and effectively infinite degrees of freedom, their state representations are inherently high-dimensional and must be discretized or encoded at increasing resolutions. 
    The demonstrated scalability of \textsc{SoftAE} in such settings underscores its potential for deployment on soft robots with rich sensory feedback or fine-grained state estimation.
    Together, these results \breview{show that} \textsc{SoftAE} \breview{enables} autonomous and scalable dynamics learning in soft robotics across platforms \breview{and} tasks, \breview{including on} real-world \breview{hardware}.

    % \subsection*{Possible Extensions}
    While the framework \breview{performs well across the diverse systems studied}, it also \breview{has limitations that motivate further work}.
    First, although the method scales efficiently to moderately high-dimensional state spaces, its performance with very high-dimensional sensory modalities, such as vision or tactile inputs, remains to be investigated. 
    We anticipate that extending active exploration to raw sensory inputs can be achieved by combining epistemic uncertainty with learned latent representations, as demonstrated in \cite{sekar2020planning, sukhija2025optimism}, which could enhance robustness and applicability in unstructured environments.
    % Second, the current implementation assumes access to relatively accurate state measurements. In future work, integrating multimodal sensing (e.g., RGB-D, tactile, or proprioception with noise models) could improve robustness and applicability in unstructured environments.
    
    \review{A further limitation is that \textsc{SoftAE} currently treats coverage of the reachable state--action space as uniformly valuable. This is well aligned with learning reusable, task-agnostic dynamics models, but it may become inefficient in very high-dimensional systems where only a lower-dimensional subset of states is relevant for the downstream task family. A promising direction is therefore to combine uncertainty-driven exploration with task-, preference-, or semantics-aware relevance signals to preserve the benefits of active dynamics learning while avoiding excessive exploration of dynamically irrelevant regions. For example, \cite{sukhija2025optimism} balances epistemic uncertainty with extrinsic task reward. More broadly, recent approaches such as \cite{escontrela2023videopredictionmodelsrewards,sancaktar2025senseisemanticexplorationguided} show how learned priors from videos or foundation-model feedback can define behaviorally meaningful reward signals that direct exploration toward relevant regions of the environment.}

    Additionally, \textsc{SoftAE} currently operates in episodic settings with fixed-length exploration budgets and offline model learning. Prior works have theoretically and empirically shown that uncertainty-based exploration can also be extended to the non-episodic setting \cite{sukhija2024neorl, sukhija2025optimism}.
    Enabling closed-loop, online model refinement during long-horizon deployments would further improve adaptability. 
    Another promising direction is to integrate low-fidelity physical models as priors, which has been shown to yield orders-of-magnitude improvements in sample efficiency \cite{rothfuss2024bridging}.
    
    One practical limitation of the current implementation is the use of iCEM optimizer for model predictive control. 
    While sample-efficient, it can be computationally demanding and less responsive in high-frequency control scenarios. 
    For tasks involving sparse rewards, long-term planning horizons, or rapid adaptation to dynamic environments, an alternative is to integrate \textsc{SoftAE} with model-based reinforcement learning (MBRL)~\cite{hafner2019dream, janner2019trust, iten2025scalable}.
    In such hybrid frameworks, simulated rollouts from the learned dynamics model are used to train a policy while uncertainty-aware exploration continues to guide real-world data collection.
    However, MBRL can suffer from instability due to compounding model errors, making it less reliable as a general replacement for trajectory optimization. 
    To demonstrate the consistency of our approach across different control paradigms, we evaluate a model-based policy optimization variant of \textsc{SoftAE} in Supplementary \format{Methods}. 
    The results show performance comparable to trajectory optimization and confirm that active exploration yields broader coverage and better generalization than \textsc{Random} exploration or task-specific learning.

    Beyond the context of soft robotics, these findings illustrate that uncertainty-guided active exploration offers a general strategy for data-efficient model learning in nonlinear, high-dimensional physical systems where analytical modeling and exhaustive data collection are impractical.
    % Applied across distinct morphologies \review{and contact regimes}, \textsc{SoftAE} points toward unified frameworks for embodied intelligence that capture the rich and diverse dynamics of real-world physical systems.

\section*{\format{Methods}}

\subsection*{Active Exploration of State-Space Models}
We study a general discrete-time nonlinear dynamical system of the form $\bm{s}_{t+1} = \bm{f}^*(\bm{s}_t, \bm{a}_t) + \bm{w}_t$, where $\vs_t \in \setS \subseteq \R^{d_\vs}$ is the state, $\bm{a}_t \in \mathcal{A} \subseteq \R^{d_\va}$ the control action, and $\vw_t \in \setW \subseteq \R^{d_\vs}$ the process noise. The system dynamics $\vf^*$ are unknown and we aim to learn them from data.  To this end, we consider the episodic \review{interaction} setting with episodes $n \in \setmath{1, \ldots, N}$.
At the beginning of episode $n$, we select and roll out a policy $\vpi_n: \setS \mapsto \setA$ for a horizon of $T$ steps on the true system.  
% We then use the data collected from the rollout $\vtau^{\vpi_n}$ to learn an estimate $\vmu_n$ of $\vf^*$. 
We then use the data collected from the rollout $\vtau^{\vpi_n}$ to learn an estimate $\vmu_n$ of $\vf^*$\review{, where $\vmu_n$ denotes the mean prediction of the learned dynamics model after $n$ episodes}.
However, when learning an unknown function, simply obtaining a single estimate is often insufficient since it does not quantify our lack of knowledge about the true function. In particular, in areas of the state--action space where we have limited data, we would expect our estimate to be less accurate compared to areas where we have collected data in abundance. 

To capture our confidence about $\vmu_n$, we additionally estimate the uncertainty $\vsigma_n$ around its predictions. Intuitively, in regions where we have less data, we would expect the uncertainty to be high, and low in regions where we have collected lots of data.
There are several uncertainty estimation models that can be used for this purpose, the most classical one being Gaussian process regression~\cite{rasmussen2005gp}. However, Bayesian deep learning models, in particular ensembles, are also often used for uncertainty quantification~\cite{lakshminarayanan2017simple}. Similar to prior work~\cite{chua2018deep, sekar2020planning, sukhija2024optimistic}, we use probabilistic ensembles for uncertainty quantification in this work. 

Thus far we have discussed how we leverage the data collected from the rollouts to learn an uncertainty-aware model of the underlying dynamics. In the following, we discuss how the policy $\vpi_n$ is selected during each episode for data collection. The goal of our algorithm is to approximate $\vf^*$ over the reachable state--action space. 
To this end, we study an active exploration setting using an intrinsic \breview{exploration} reward function \breview{$r^{\mathrm{exp}}_n$} \breview{for episode $n$} based on the epistemic uncertainty about $\vf^*$ \breview{that remains after the first $n-1$ episodes,} as advocated by Sukhija \textit{et al.}~\cite{sukhija2024optimistic, sukhija2025optimism}\breview{,}
\begin{align}
    \breview{
    r^{\mathrm{exp}}_n(\vs_t,\va_t)
    =
    \norm{\vsigma_{n-1}(\vs_t,\va_t)}.
    }
    \label{eq:exploration_reward}
\end{align}

To efficiently explore the system dynamics, 
% one tempting approach would be to optimize for the following objective:
one tempting approach would be to \breview{select the policy that maximizes the expected sum of intrinsic rewards $r_n^{\mathrm{exp}}$ over an episode}:
  \begin{align}
      \vpi^*_n = \argmax_{\vpi \in \Pi}& \hspace{0.2em} J_n(\vpi) = \argmax_{\vpi \in \Pi} \hspace{0.2em} \E_{\vtau^{\vpi}}\left[\sum_{t=0}^{T-1} \norm{\vsigma_{n-1}(\vs_t, \vpi(\vs_t))}\right],       \label{eq:exploration_op} \\
      \vs_{t+1} &= \vmu_{n-1}(\vs_t, \vpi(\vs_t)) + \vw_t. \label{eq:exploration_op_dyn}
  \end{align}
The \breview{intrinsic} reward in \cref{eq:exploration_op} encourages the agent to explore regions with high uncertainty, since high uncertainty regions typically correspond to regions where we have less data. Thus, by maximizing the uncertainty of our model, we encourage the agent to efficiently cover the state--action space. To obtain a policy, any policy or trajectory optimization technique can be used. \review{In our implementation, we instantiate this optimization using receding-horizon trajectory optimization with model-predictive control (MPC) or model-based policy optimization (MBPO).} Once we have learned the underlying dynamics well, we can leverage our model $\left(\vmu_n,\vsigma_n\right)$ for analyzing and controlling the system. Moreover, given any reward function, we can solve the underlying task by using our learned model for planning.

Observe that in \cref{eq:exploration_op_dyn} above, we use the mean model $\vmu_n$ for planning the state propagation. Combined with the active exploration objective, we call this algorithm \textsc{Mean-AE}. Buisson-Fenet \textit{et al.} also propose a similar approach~\cite{buisson2020actively}. However, Chua \textit{et al.} found that planning with the mean often underperforms in practice since the policy exploits the inaccuracies in the mean model~\cite{chua2018deep}. Instead, they propose a trajectory sampling approach for uncertainty propagation, which we adopt as follows:
 \begin{align}
      \vpi^*_n = \argmax_{\vpi \in \Pi}& \hspace{0.2em} J_n(\vpi) = \argmax_{\vpi \in \Pi} \hspace{0.2em} \E_{\vtau^{\vpi}}\left[\sum_{t=0}^{T-1} \norm{\vsigma_{n-1}(\vs_t, \vpi(\vs_t))}\right],       \label{eq:exploration_ts} \\
      \vs_{t+1} &\sim \setN\left(\vmu_{n-1}(\vs_t, \vpi(\vs_t)), \vsigma^2_{n-1}(\vs_t, \vpi(\vs_t))\right) + \vw_t. \label{eq:exploration_ts_dyn}
  \end{align}
  Similar to domain randomization, by sampling proportional to the model uncertainty in \cref{eq:exploration_ts_dyn}, we make our policy robust towards model inaccuracies while also optimizing for the active objective in \cref{eq:exploration_ts}. We call this algorithm \textsc{PETS-AE}. 
    \review{However, because \textsc{PETS-AE} optimizes expected performance under sampled dynamics, it can miss regions that are highly informative but reachable only under a small subset of plausible dynamics.}

\review{
To make this contrast precise, consider a finite-model setting in which, at episode $n$, the learner maintains a set of plausible dynamics models \(\mathcal{M}_n=\{\vf_1,\ldots,\vf_M\}\), with \(\vf^*\in\mathcal{M}_n\). Let \(J_n^{\mathrm{exp}}(\vpi,\vf)\) denote the exploration objective obtained by rolling out policy \(\vpi\) under model \(\vf\). 
Then, \textsc{PETS-AE} can be viewed as selecting a policy that performs well on average across plausible dynamics sampled from a distribution \(p_n\) supported on \(\mathcal M_n\):}
 \begin{align}
    \review{
    \vpi_n^{\mathrm{PETS\text{-}AE}}
    =
    \arg\max_{\vpi\in\Pi}
    \mathbb{E}_{\vf\sim p_n}
    \left[
        J_n^{\mathrm{exp}}(\vpi,\vf)
    \right].
    }
\end{align}
\review{
By contrast, we propose to follow the \emph{optimism in the face of uncertainty} principle: instead of averaging over plausible dynamics, we select the most favorable plausible dynamics model for the exploration objective, which we call \textsc{SoftAE}:}
\begin{align}
    \review{
    \vpi_n^{\textsc{SoftAE}}
    =
    \arg\max_{\vpi\in\Pi}
    \max_{\vf\in\mathcal{M}_n}
    J_n^{\mathrm{exp}}(\vpi,\vf).
    }
\end{align}
\review{
Thus, \textsc{PETS-AE} favors policies that are informative on average across sampled dynamics, whereas \textsc{SoftAE} favors policies that could reach highly informative regions under at least one plausible dynamics realization. This turns epistemic uncertainty into directed exploration. In our setting, the plausible model set is continuous and represented by the confidence set $\mathcal{M}_n$ induced by the learned mean model of the previous episode \(\vmu_{n-1}\), epistemic uncertainty \(\vsigma_{n-1}\), and a scaling parameter \(\beta_{n-1}\). Under suitable regularity assumptions, $\beta_{n-1}$ can be chosen according to theoretical confidence bounds such that the true dynamics lie in $\mathcal{M}_n$ with high probability~\cite{chowdhury_kernelized_2017,rothfuss_hallucinated_2023}. Consequently, the optimistic objective upper bounds the exploration objective under the true dynamics,}
\begin{align}
    \review{
    \max_{\vf\in\mathcal{M}_n}
    J_n^{\mathrm{exp}}(\vpi,\vf)
    \geq
    J_n^{\mathrm{exp}}(\vpi,\vf^*),
    \qquad
    \forall \vpi\in\Pi .
    }
\end{align}
\review{
Following the optimistic planning formulation of Curi \textit{et al.}~\cite{curi2020efficient}, we implement the maximization over plausible dynamics through an auxiliary uncertainty variable \(\veta\). We adapt their optimistic planner as follows:   }
   \begin{align}
      \vpi^*_n = \argmax_{\vpi \in \Pi, \veta \in [-\beta, \beta]^{d_\vs}}& \hspace{0.2em} J_n(\vpi) = \argmax_{\vpi \in \Pi, \veta \in [-\beta, \beta]^{d_\vs}} \hspace{0.2em} \E_{\vtau^{\vpi}}\left[\sum_{t=0}^{T-1} \norm{\vsigma_{n-1}(\vs_t, \vpi(\vs_t))}\right],       \label{eq:exploration_optimistic} \\
      \vs_{t+1} &= \vmu_{n-1}(\vs_t, \vpi(\vs_t)) + \vsigma_{n-1}(\vs_t, \vpi(\vs_t)) \veta(\vs_t) + \vw_t. 
    \label{eq:exploration_optimistic_dyn}
  \end{align}
  \review{The auxiliary variable \(\veta\) parameterizes the choice of plausible dynamics within the continuous confidence set}
\begin{align}
    \review{
    \setM(\vs_t) = [ \vmu_{n-1}(\vs_t, \va_t) \pm \beta_{n-1} \vsigma_{n-1}(\vs_t, \va_t)],
    }
\end{align}
\review{where \(\beta_{n-1}\) controls the size of the confidence set. While $\beta_{n-1}$ admits a theoretically motivated choice, we treat it as a hyperparameter in practice, as is common in optimistic and intrinsic exploration methods where the amount of exploration is selected heuristically~\cite{kakade_information_2020, curi2020efficient}. Importantly, \(\veta\) is not an additional physical control input: the robot still executes only \(\va_t=\vpi(\vs_t)\). Instead, \(\veta\) selects the favorable dynamics realization used for optimistic planning.} We call this algorithm \textsc{SoftAE}, as summarized in \cref{algorithm:SoftAE}.

\review{
This mechanism is particularly useful for soft robots, where informative states often correspond to large deformations, contact-rich configurations, swimming gaits, or hopping motions that require coordinated multi-step actuation. Such behaviors can be missed by random exploration and by posterior-sampling planners that optimize average performance over plausible dynamics. By planning optimistically through the uncertainty set, \textsc{SoftAE} induces additional directed exploration toward hard-to-reach but informative regimes.
}

  Sukhija \textit{et al.} propose the planning problem in \cref{eq:exploration_optimistic,eq:exploration_optimistic_dyn} and show that under common regularity assumptions on $\vf^*$, the objective guarantees polynomial sample complexity~\cite{sukhija2024optimistic}, i.e., that we converge to an $\epsilon$-optimal solution in polynomial time. 
  This in turn implies that the estimate $\vmu_n$ converges to the true system $\vf^*$ for $N\rightarrow\infty$. 
  Recently, Sukhija \textit{et al.} show that convergence also holds for both \textsc{Mean-AE} and \textsc{PETS-AE}~\cite{sukhija2025optimism}. 
  However, while all the aforementioned planning strategies guarantee convergence in theory, \review{optimism-based methods yield tighter theoretical bounds because the maximization over the confidence set induces additional directed exploration rather than relying on sampled futures to reveal informative trajectories. Empirically, this advantage carries over to the soft robotic tasks: across all environments, \textsc{SoftAE} attains the most consistent and highest returns among the three active exploration strategies. We report the full comparison between \textsc{Mean-AE}, \textsc{PETS-AE}, and \textsc{SoftAE} in \format{Supplementary Fig.}~\ref{figS:2_ae_comparison}.}
  
  \review{To further investigate the practical consequences of this distinction, we additionally consider exploration under an actuation penalty. Specifically, we augment the intrinsic exploration reward} \breview{in \cref{eq:exploration_reward} with a quadratic action cost,}  
\begin{align}
    \breview{
    r_n^{\mathrm{exp,act}}(\vs_t, \va_t) = 
    r_n^{\mathrm{exp}}(\vs_t, \va_t) - c_{\mathrm{act}}\norm{\va_t}_2^2}
    = \norm{\vsigma_{n-1}(\vs_t,\va_t)} - c_{\mathrm{act}}\norm{\va_t}_2^2,
\end{align}

\review{where \(c_{\mathrm{act}}\) controls the trade-off between information gain and actuation effort.} 
% \breview{We use $r_n^{\mathrm{exp,act}}$ in place of $r_n^{\mathrm{exp}}$ in all three exploration objectives, \cref{eq:exploration_op,eq:exploration_ts,eq:exploration_optimistic}.}
\breview{In all three exploration objectives, \cref{eq:exploration_op,eq:exploration_ts,eq:exploration_optimistic}, the $r_n^{\mathrm{exp}}$ term $\norm{\vsigma_{n-1}(\vs_t, \vpi(\vs_t))}$ is then replaced by $r_n^{\mathrm{exp,act}}(\vs_t, \vpi(\vs_t))$.}
\review{This setting exposes a regime in which the difference between optimistic planning and average-case posterior sampling becomes practically important: when exploration must seek informative states while limiting actuation effort, \textsc{SoftAE} consistently outperforms \textsc{Mean-AE} and \textsc{PETS-AE}, whereas planning with the mean or sampled dynamics tends to favor conservative trajectories. The full sweep over \(c_{\mathrm{act}}\in\{0,0.25,0.5,0.75,1.0\}\) and the accompanying analysis are reported in Supplementary \format{Fig.}~\ref{figS:3_action_cost}.}

% \newpage
% \begin{algorithm}[ht]
%     \caption{\textbf{\textsc{SoftAE}}}
%     \label{algorithm:SoftAE}
%     \begin{algorithmic}[]
%         \STATE {\textbf{Init:}}{ Aleatoric uncertainty $\sigma$, Statistical model $(\vmu_0, \vsigma_0, \beta_0)$, $\setD_0 \coloneqq \emptyset$}
%         \FOR{episode $n=1, \ldots, N$}{
%             \vspace{-0.5cm}
%             \STATE {
%             \begin{align*}
%                 &\vpi_n =  \textsc{OptimizePolicy}(\vmu_n, \vsigma_n, \beta_n)  &&\text{\ding{228} Prepare policy via Equations (\ref{eq:exploration_optimistic}--\ref{eq:exploration_optimistic_dyn})} \\
%                 &(\mS, \mA, \mS') \leftarrow \textsc{Rollout}(\vpi_n) \quad &&\text{\ding{228} Collect measurements } \\
%                  &\setD_n \leftarrow \setD_{n-1} \cup (\mS, \mA, \mS') &&\text{\ding{228} Add rollout to dataset}\\
%                  &\text{Update } (\vmu_n, \vsigma_n, \beta_n, \setD_{n}) \quad &&\text{\ding{228} Update model via \cref{eq:fsvgd_updates}}
%             \end{align*}
%             }}
%         \ENDFOR
%     \end{algorithmic}
% \end{algorithm}
\begin{algorithm}[t]
    \caption{
        \textbf{\textsc{SoftAE}\review{: Soft-robot Active Exploration.}}
    }
    \label{algorithm:SoftAE}
    \begin{algorithmic}[]
        \STATE \textbf{Init:}
        Statistical dynamics model
        $(\vmu_0,\vsigma_0,\beta_0)$ and dataset
        $\setD_0 \coloneqq \emptyset$

        \FOR{episode $n=1,\ldots,N$}
            \vspace{-.7cm}
            \STATE {
\begin{alignat*}{2}
    &\review{(\vpi_n,\veta_n)
    \in
    \argmax_{\substack{
        \vpi\in\Pi,\\
        \veta(\cdot)\in
        [-\beta_{n-1},\beta_{n-1}]^{d_{\vs}}
    }}
    \mathbb{E}\bigg[
        \sum_{t=0}^{T-1}
        \big\|
        \vsigma_{n-1}}
        &&\review{\bigl(
            \widetilde{\vs}_t,
            \vpi(\widetilde{\vs}_t)
        \bigr)
        \big\|
    \bigg]}
    \\
    &&&
        \text{\ding{228} \review{Prepare optimistic policy via
        \eqref{eq:exploration_optimistic}}}
    \\[0.2em]
    &\quad\text{\review{s.t.}}\quad\,\,\,\,\,\
    \review{\widetilde{\vs}_{t+1}
    =
    \vmu_{n-1}
    \bigl(
        \widetilde{\vs}_t,
        \vpi(\widetilde{\vs}_t)
    \bigr)
    +    
    \vsigma_{n-1}}
    &&\review{\big(
        \widetilde{\vs}_t,
        \vpi(\widetilde{\vs}_t)
    \big)
    \veta(\widetilde{\vs}_t)}
    \\
    &&&
        \text{\ding{228} \review{Propagate optimistic dynamics via
        \eqref{eq:exploration_optimistic_dyn}}}
    \\[0.2em]
    &(\mS,\mA,\mS')
    \leftarrow
    \textsc{Rollout}(\vpi_n)
    &&
        \text{\ding{228} Collect measurements}
    \\
    &\setD_n
    \leftarrow
    \setD_{n-1}\cup(\mS,\mA,\mS')
    &&
        \text{\ding{228} Add rollout to dataset}
    \\
    &(\vmu_n,\vsigma_n,\beta_n)
    \leftarrow
    \textsc{UpdateModel}(\setD_n)
    &&
        \text{\ding{228} Update model via \eqref{eq:fsvgd_updates}}
\end{alignat*}

            }
        \ENDFOR
    \end{algorithmic}
\end{algorithm}

\subsection*{Learning Uncertainty Aware Dynamics}
\review{
To instantiate the confidence set \(\mathcal{M}_n\) used by \textsc{PETS-AE} and \textsc{SoftAE}, we require a learned mean dynamics model \(\vmu_n\) together with an epistemic uncertainty estimate \(\vsigma_n\). We obtain both quantities using an ensemble of neural networks (NNs)~\cite{lakshminarayanan2017simple}. The individual ensemble members represent plausible dynamics models consistent with the data collected so far, and their disagreement provides the uncertainty estimate used for active exploration.
}
We approximate the posterior distribution with a set of $L$ NN parameter particles $\{ \vtheta_1, \dots, \vtheta_L \}$. The particles are sampled i.i.d. from a parameter distribution $\vtheta^0_l \sim p(\vtheta)$ and 
updated via maximum-likelihood estimation (MLE):
\begin{equation} \label{eq:fsvgd_updates}
    \vtheta_l^{i+1} \leftarrow \vtheta_l^i + \underbrace{\gamma}_{\text{Learning Rate}}  \underbrace{ \nabla_{\vtheta_l^i} \ln p(\mS'|\mS,\mA, \vtheta_l^i)}_{\text{MLE} \vspace{-4pt}}~,\forall l \in \{1, \dots, L\}.
\end{equation}
We pick $p(\vs'|\vs,\va, \vtheta_l^i) \coloneqq \setN(\textbf{NN}(\vs, \va| \vtheta_l^i), \epsilon^2)$, resulting in the squared error loss for each particle. 

Given the particles $\{ \vtheta^n_1, \dots, \vtheta^n_L \}$ at iteration $n$, we estimate the mean and model uncertainty with
\begin{align}
    \vmu_n(\vs, \va) \approx \frac{1}{L}\sum_{i=1}^L \textbf{NN}(\vs, \va|\vtheta^n_i), \quad
    \vsigma^2_n(\vs, \va) \approx \text{Var}\left(\{\textbf{NN}(\vs, \va|\vtheta^n_i)\}^L_{i=1}\right)
\end{align}

Intuitively, as we collect more data in our domain $\setS \times \setA$, we expect the different particle initializations to converge to the same prediction and therefore our epistemic uncertainty to decrease. This holds for classical models such as Gaussian processes~\cite{rasmussen2005gp} and has also been shown empirically for NN models~\cite{pathak2019self}.

\subsection*{\review{Practical} Trajectory and Policy Optimization \review{ for Exploration and Downstream Control}}

\review{
The objectives (\ref{eq:exploration_op}--\ref{eq:exploration_optimistic_dyn}) above define the exploration problem at the level of a policy optimization problem over \(\vpi\). In practice, we do not optimize an arbitrary continuous policy function directly. Instead, we instantiate the optimization in two ways: receding-horizon trajectory optimization with model-predictive control (MPC), and model-based policy optimization (MBPO). Both approaches use the learned dynamics model $\vmu_n$ to evaluate candidate future trajectories, but differ in how the controller is represented. In MPC, the controller is implicit: at each time step, we optimize a finite sequence of actions, execute only the first, observe the next state, and replan. In MBPO, the controller is explicit: a parametrized policy is trained on real data augmented with short imagined rollouts from the learned dynamics model.
}

\review{
The same control method is used across two phases of our pipeline. During \emph{active exploration}, candidate trajectories are scored with the intrinsic exploration reward \(\breview{r^{\mathrm{exp}}_n=}\norm{\vsigma_{n-1}(\vs_t,\va_t)}\), and in \textsc{SoftAE} they are propagated through the optimistic dynamics in \cref{eq:exploration_optimistic_dyn}. During \emph{downstream task evaluation}, the same optimization is reused with task-specific rewards \(r_t(\vs,\va)\) defined for each environment in \format{Supplementary} Table~\ref{tabS:1_rewards}. 
For \textsc{SoftAE} and \textsc{Random}, downstream task rollouts are used only for evaluation and are not added to the training dataset, and no task-specific fine-tuning of the dynamics model is performed. 
Exploration and task execution are therefore fully separated: the model is learned from task-agnostic exploration data, and task rewards are applied afterward to test whether this learned model supports zero-shot control. The return-over-episode curves are obtained by evaluating the current learned model at successive exploration checkpoints using this task-centric planner. In contrast, the \textsc{H-UCRL} baseline uses the task reward during training and therefore represents a task-specific model-based baseline. It is trained on a subset of the tasks (\textit{(i)}, \textit{(iii)}, \textit{(v)}, and \textit{(vii)}) while the remaining tasks are held out as zero-shot generalization tests.
}

\review{Our main experiments use MPC with iCEM~\cite{pinneri2021sample} for both exploration and downstream control. To demonstrate that the framework is not tied to a specific control paradigm, we additionally evaluate an MBPO variant of \textsc{SoftAE} (\format{Supplementary Fig.}~\ref{figS:4_mbpo}). The two methods are detailed below.}

\paragraph{Trajectory Optimization and Model-Predictive Control} Instead of learning an explicit policy function, here, we directly optimize over the actions $\{\va_0, \va_1, \dots\}$ for a fixed horizon $H < T$ and apply receding horizon/model predictive control~\cite{morari1999model}. For instance, instead of solving for the policy in \cref{eq:exploration_op},  we solve the following trajectory optimization problem:
  \begin{align}
     \{\va^*_0, \va^*_1, \dots, \va^*_{H}\}  &= \argmax_{\va_0, \va_1, \dots, \va_H \in \setA^{H}} \hspace{0.2em} J_n(\{\va_0, \va_1, \dots, \va_H\} ) \label{eq:exploration_op_mpc} \\
     &= \argmax_{\va_0, \va_1, \dots, \va_H \in \setA^{H}} \hspace{0.2em} \E_{\vtau^{\vpi}, \hat{\vs}_0 = \vs_t}\left[\sum_{h=0}^{H-1} \norm{\vsigma_{n-1}(\hat{\vs}_h, \va_h)}\right],       \notag \\
      \hat{\vs}_{h+1} &= \vmu_{n-1}(\hat{\vs}_h, \va_h) + \vw_t. \notag
  \end{align}
  Next, we apply the first control $\va^*_0$ to the real system and observe the next state $\vs_{t+1}$. We repeat the optimization above with $\hat{\vs}_0 = \vs_{t+1}$ to obtain the next control input. By repeating the optimization at each timestep, we obtain an implicit closed-loop controller. This approach does not require a policy parameterization, is simple, and is often more stable during learning. We use a sampling-based solver, in particular \review{the improved cross-entropy method iCEM} \cite{pinneri2021sample}, for \cref{eq:exploration_op_mpc}. Here, we use a candidate distribution over the action sequence $p(\{\va_0, \va_1, \dots, \va_H\})$, e.g., a clipped Gaussian, and then generate $P$ samples $\{\va^p_0, \va^p_1, \dots, \va^p_H\}^P_{p=0}$. We evaluate the objective for all the samples $J_n(\{\va^p_0, \va^p_1, \dots, \va^p_H\} )$ and use its value to update the sampling distribution $p(\{\va_0, \va_1, \dots, \va_H\})$. This procedure is repeated over several steps, following which the best candidate is returned (see \cite{pinneri2021sample} for more details).

\review{
This procedure is related to model predictive path integral control (MPPI,~\cite{mppi}) in the broad sense that both are sampling-based MPC methods: they sample candidate action sequences, roll them out through a dynamics model, score the resulting trajectories, and update the sampling distribution before executing the first action in a receding-horizon manner. However, our implementation uses iCEM rather than MPPI. While MPPI updates the control distribution using path-integral importance weights over all sampled trajectories, iCEM follows a cross-entropy method update based on elite action sequences.}

\paragraph{Model-Based Policy Optimization}
As an alternative to trajectory optimization, we can also learn a parametrized policy $\vpi_{\vphi}$ directly using the learned dynamics. Concretely, we sample a batch of $P$ states $\{\vs_i\}_{i=1}^P$ from a replay buffer of real state transitions gathered so far and train the policy with Soft Actor-Critic (SAC~\cite{haarnoja2018sac}), which combines off-policy learning with entropy maximization to balance exploitation and exploration. We augment the limited real-world data with synthetic rollouts of length $H$ generated from the learned dynamics. For each sampled state, we simulate
\begin{align}
\hat{\vs}_{h+1} &= \vmu_{n-1}(\hat{\vs}_h, \va_h) + \vw_h, \quad \va_h \sim \vpi_{\vphi}(\cdot|\hat{\vs}_h), \quad h=0,\dots,H-1,
\end{align}
with $\hat{\vs}_0 = \vs_i$. The resulting imagined trajectories ${(\hat{\vs}_h, \va_h, \hat{\vs}_{h+1})}$ branched from real data are then combined with real transitions to form the training set for policy optimization, which improves sample efficiency \cite{janner2019trust}. To further stabilize learning, we apply the symlog transform~\cite{hafner2019dream} to observations in order to reduce sensitivity to outliers. Specifically, for a scalar input $x$ we train the dynamics model on
\begin{align}
    \operatorname{symlog}(x) = \operatorname{sign}(x)\log\left(1+|x|\right),
\end{align}
which preserves the sign and compresses large magnitudes while remaining approximately linear near zero. Its inverse for mapping back to the original scale is
\begin{align}
    \operatorname{symexp}(y) = \operatorname{sign}(y)\left(e^{|y|}-1\right),
\end{align}
which yields bounded, well-behaved gradients and helps prevent rare spikes in sensor values from destabilizing training.

Unlike trajectory optimization, which repeatedly solves an open-loop action sequence optimization at each time step, model-based policy optimization amortizes this computation into the policy parameters $\vphi$. This yields a closed-loop controller that is both sample-efficient---thanks to model rollouts---and computationally efficient at deployment. However, a potential drawback is that inaccuracies in the learned dynamics can mislead the policy, especially early on. If $\vpi_{\vphi}$ then exploits model errors, this bias can propagate through imagined rollouts and destabilize subsequent learning.

    \subsection*{Soft Robotic Simulated Environments and Tasks}
        \subsubsection*{Soft Continuum Arm via Cosserat Rod Simulation}
            The simulated soft robotic arm environment is grounded in Cosserat rod theory, implemented using the Elastica simulator \cite{Naughton2021} \review{with an integration timestep of $1 \times 10^{-4}$\,s} (\format{Supplementary Movie~2}). 
            The system models a slender continuous soft arm that exhibits nonlinear bending deformation. 
            Actuation is applied as internal torques in the normal and binormal planes, controlled via 6 evenly spaced control points along the arm. 
            This results in a total of 12 actuation degrees of freedom. 
            The system’s 58-dimensional state space includes spatially distributed representations of position, linear velocity, orientation, and angular velocity along the length of the continuously deformable arm (\format{Fig.}~\ref{fig:1_overview}A), making the dynamics both high-dimensional and highly nonlinear. 
            \review{A control frequency of 100\,Hz is used for both exploration and downstream control.}

            We evaluate the learned dynamics models on two endpoint reaching tasks that differ in difficulty. In both cases, the goal is to control the soft arm to reach a fixed target position within its workspace. Task \textit{(i)} places the target at a small distance away from the rest state, requiring only moderate deformation and resulting in relatively smooth, low-curvature trajectories. The second task \textit{(ii)} places the target closer to the edge of the workspace, demanding more extreme bending and invoking stronger nonlinear dynamics. 
        
        \subsubsection*{Articulated Fish with Deformable Skin in Fluid}
            The second simulated system is a soft robotic fish in a fluid environment, implemented using FishGym~\cite{liu2022fishgym}, a high-performance simulator with two-way coupled fluid-structure interaction\review{, run with an integration timestep of $4 \times 10^{-3}$\,s} (\format{Supplementary Movie~3}).
            As illustrated in \format{Fig.}~\ref{fig:1_overview}B, the robot consists of an articulated skeleton enclosed in a deformable skin mesh and is immersed in a simulated viscous fluid.
            This setup captures complex hydrodynamic effects such as vortex shedding, body--fluid coupling, and delayed actuation responses, making it a challenging environment for dynamics learning.

            The fish is actuated through four internal joints along its spine, each controlled by a scalar torque input, enabling flexible body undulation for thrust generation. \review{A control frequency of 10\,Hz is used for both exploration and downstream control, reflecting the slower hydrodynamic timescales of fluid--structure interaction.}
            To assess the generalization of the learned dynamics model, we define four aquatic locomotion tasks: \textit{(iii)} swim forward along the +x direction, \textit{(iv)} execute a U-turn and swim along the –x direction, \textit{(v)} reach a randomized target location, and \textit{(vi)} swim away from a randomized target. 
            For all tasks, we constrain the robot fish to swim on a horizontal 2D plane without buoyancy control.
            Tasks (iii) and (iv) use a 15-dimensional state space comprising the fish’s velocity, orientation, and joint positions and velocities. 
            Tasks (v) and (vi) extend this with an additional 3-dimensional vector representing the relative target position to the fish. 

        \subsubsection*{Musculoskeletal Leg with Electrohydraulic Muscles}
            Inspired by recent advances in electrohydraulic musculoskeletal legs~\cite{buchner2024electrohydraulic}, the third simulated environment models a planar robotic leg capable of fast, adaptive motion with tunable stiffness and high energy efficiency (\format{Supplementary Movie~4}). 
            The leg is rigidly attached to a circular boom that rotates around a fixed vertical axis, constraining the system to move along a horizontal arc. 
            Actuation is provided at the hip and knee joints via antagonistic pairs of contracting electrohydraulic artificial muscles, enabling compliant, muscle-driven dynamics.
            
            To simulate this system, we implement the musculoskeletal (MSK) leg in MuJoCo (\format{Fig.}~\ref{fig:1_overview}C) \review{with an integration timestep of $2 \times 10^{-3}$\,s}. The soft muscle actuators are modeled using a neural network trained on real electrohydraulic actuator data, capturing nonlinear voltage-to-force characteristics. 
            An additional DC motor added at the boom joint allows us to control the boom’s roll angle, which effectively adjusts the pitch and hopping direction of the leg. This motor enables hopping behaviors in different directions on a circular track.

            The state space is 10-dimensional and includes the yaw, pitch, and roll angles of the boom, the hip and knee joint angles, and the corresponding angular velocities for each of these five degrees of freedom.
            The action space comprises 4 muscle voltage commands and the position control signal to the DC motor. 
            \review{A control frequency of 100\,Hz is used for both exploration and downstream control.}
            We define two locomotion tasks for this system: \textit{(vii)} hop counterclockwise and \textit{(viii)} hop clockwise along the circular trajectory.
            These tasks require coordinated actuation between the compliant leg and boom orientation, highlighting the challenge of learning hybrid, muscle-driven dynamics.

\section*{\review{Data Availability}}
\breview{Data sharing not applicable to this article as no datasets were generated or analysed during the current study. }
\review{The source data underlying all figures in the main text and Supplementary Information are provided with this paper. Other data supporting the findings of this study can be regenerated from the released code (see Code Availability).}

\section*{\review{Code Availability}}
The code accompanying this manuscript is \breview{available at \url{https://github.com/srl-ethz/soft_robot_active_exploration} under the MIT License, and an
archived snapshot of the version used here is deposited at Zenodo~\cite{zheng2026softae_code}. No restrictions apply to
its use or reuse.}
The release includes pretrained model checkpoints, an evaluation demo that reproduces the soft-arm reaching results in approximately ten minutes on a single \breview{NVIDIA GeForce RTX 4090} GPU, and a pre-built Docker image with all dependencies for reproducible execution. 

% \breview{During review, the repository and the Zenodo record are private. 
% We provide the code as a supplementary archive for reviewer assessment.
% An access link to the Zenodo deposit is provided to the editor via the Manuscript Tracking System. 
% Both will be made publicly available upon acceptance.}

\bibliographystyle{sn-nature-fixed-italic}
\bibliography{references}

\textbf{Acknowledgements:}~This work was funded, in part, by a generous donation from UBS to the ETH Foundation to create a new chair for Robotics at ETH Zurich. This work also received funding from the Swiss National Science Foundation under NCCR Automation, grant agreement 51NF40 180545. H.Z. and C.L. were partially funded by a doctoral fellowship from the ETH AI Center. H.Z. was also partially supported by the SDSC Grant entitled `C22-08: Data-Driven Inference of Mesh-based Representations for Deformable Objects from Unstructured Point Clouds'. B.S. was supported by ELSA (European Lighthouse on Secure and Safe AI) funded by the European Union under grant agreement No. 101070617. 
We would like to thank L. Treven for initial discussions and assistance with algorithm visualizations, W. Song for providing access to FishGym, T. Buchner for help in adapting the electrohydraulic musculoskeletal robotic leg and its visualization, A. Kazemipour for support in parsing the MuJoCo simulation XML configuration for the leg, and J. Cheng for help with the SAC model-based variant. \\\\

\textbf{Author Contributions:}~H.Z., B.S., A.K., and R.K. conceptualized the project. H.Z. and B.S. designed the experiments, implemented the coding framework, and analyzed the data. H.Z. conducted the simulation experiments. H.Z. and C.L. performed the real-world experiments and prepared the visualizations. K.I. implemented and conducted the model-based policy optimization experiments. A.K. and R.K. acquired funding, provided computational and hardware infrastructure, and supervised the work. All authors contributed to the manuscript and approved the final draft.\\\\

\textbf{Competing Interests}:~The authors declare no competing interests.

% \\\\
% \textbf{Data and Materials Availability:} All data needed to support the conclusions of this manuscript are included in the main text or supplementary materials.

% \section*{Supplementary Material}
%     \format{Supplementary} Methods\\
%     \format{Supplementary} Figs. \format{1} to \format{5}\\
%     \format{Supplementary} Tables \format{1} to \format{8}\\
%     \format{Supplementary} References\\
%     \format{Supplementary} Movies \format{1} to \format{5}\\
%     Code

% \clearpage
% \input{2-Figures}

%%%%%%%%%%%%%%%% END OF MAIN TEXT %%%%%%%%%%%%%%%

\newpage

%%%%%%%%%%%%%%%% START OF SUPPLEMENT %%%%%%%%%%%%%%%

% Figures, tables, equations and pages in the supplement are numbered S1, S2 etc.
% \renewcommand{\thefigure}{S\arabic{figure}}
% \renewcommand{\thetable}{S\arabic{table}}
% \renewcommand{\theequation}{S\arabic{equation}}
% \renewcommand{\thepage}{S\arabic{page}}

\makeatletter
\renewcommand{\fnum@figure}{\textbf{Supplementary Figure~\thefigure}}
\renewcommand{\fnum@table}{\textbf{Supplementary Table~\thetable}}
\makeatother

\renewcommand{\theequation}{S\arabic{equation}}
\renewcommand{\thepage}{S\arabic{page}}

\setcounter{figure}{0}
\setcounter{table}{0}
\setcounter{equation}{0}
\setcounter{page}{1} % not 0 as \newpage already started a supplementary page
% References continue the numbering from the main text.
%%%%%%%%%%%%%%%%%%%%%%%%%%%%%%%%%%%%%%%%%

\newpage
\begin{center}
% \section*{Supplementary Materials for\\ \vspace{0.1in}\scititle}
% \vspace{0.1in}
\section*{Supplementary \format{Information} for\\ Learning Soft Robotic Dynamics with Active Exploration}
\vspace{0.1in}
% Author list for the supplement
% Indicate the corresponding authors, but do NOT include institutions here
% It would be nice if the template auto-generated this, but doing so is complicated...
Hehui~Zheng,
Bhavya~Sukhija,
Chenhao~Li,\\
Klemens~Iten,
Andreas~Krause,
Robert~K.~Katzschmann$^{\ast}$\\ 
\small$^\ast$Corresponding author. Email: rkk@ethz.ch\\
\end{center}

% Fill out the numbers for each type of supplementary material,
% and delete any lines that aren't applicable.
% These are just example numbers that don't match the rest of this template.
\subsubsection*{This PDF file includes:}
Supplementary Methods\\
\format{Supplementary} Figures \format{1} to \format{5}\\
\format{Supplementary} Tables \format{1} to \breview{9}\\
Legends for Supplementary Movies 1 to 5\\
\format{Supplementary References} \\

\subsubsection*{Other Supplementary Materials for this manuscript:}
\format{Supplementary} Movies \format{1} to \format{5}\\
% Code

\newpage

%%%%%%%%%%%%%%%% MATERIALS AND METHODS %%%%%%%%%%%%%%%

\subsection*{Supplementary Methods}
    \subsubsection*{\review{Exploration Efficiency vs. Coverage: Random Exploration with Extended Budget}}
    \label{secS:random_extended}
    \review{To examine whether the performance gap between \textsc{SoftAE} and \textsc{Random} is a matter of sample efficiency alone, we extend the \textsc{Random} exploration budget on the simulated soft continuum arm from 100 to 500 episodes, five times the budget granted to \textsc{SoftAE} and \textsc{H-UCRL}.
    We then evaluate downstream performance on the same two reaching tasks \textit{(i)} and \textit{(ii)} as in \format{Fig.}~\ref{fig:2_sim_tasks}A.}
    \review{As shown in \format{Supplementary} \format{Fig.}~\ref{figS:1_random_extended}, on the easier task \textit{(i)} reach close, the extended \textsc{Random} baseline surpasses \textsc{SoftAE}'s performance at 100 exploration episodes and approaches but does not match \textsc{H-UCRL}, indicating that undirected exploration can be effective for tasks confined to a small region around the rest configuration when sufficient budget is available.
    On the harder task \textit{(ii)} reach far, however, \textsc{Random} plateaus well below \textsc{SoftAE} even with five times the interaction budget, while \textsc{H-UCRL} fails to generalize to the unseen far target. 
    This confirms that the advantage of uncertainty-driven exploration is not merely faster convergence but a qualitatively broader coverage of the state--action space, which becomes essential as task difficulty and required deformation increase.}

    \subsubsection*{Comparison of Different Uncertainty-Driven Active Exploration Strategies}
    \label{sec:additional methods}
    In the main paper, we focus on \textsc{SoftAE}, our optimistic active exploration strategy for data-efficient and task-agnostic dynamics learning in soft robotic systems.  
    Here, we compare \textsc{SoftAE} with two additional uncertainty-driven active exploration baselines, \textsc{Mean-AE} and \textsc{PETS-AE}, alongside \textsc{Random} exploration and the task-specific model-based method \textsc{H-UCRL}.

    All active exploration methods use probabilistic ensembles to model the system dynamics, providing both a mean prediction $\vmu_n$ and an uncertainty estimate $\vsigma_n$ of the next state at each state--action pair. The exploration policy $\vpi_n$ is selected to maximize predicted model uncertainty over the rollout horizon, thereby driving the agent toward less-explored regions of the state--action space. 
    \textsc{Mean-AE} plans trajectories using the mean model $\vmu_n$ directly. 
    While straightforward, this approach can lead to biased state estimate, as even small model errors can accumulate quickly over time~\cite{s-chua2018deep}.
    It is also proven suboptimal outside of linear systems~\cite{s-simchowitz2020naive}, making the mean estimator less suitable for soft robotic systems, which are inherently nonlinear and underactuated.
    \textsc{PETS-AE} follows the probabilistic ensemble trajectory sampling approach, sampling from $\mathcal{N}(\vmu_n, \vsigma_n^2)$ during planning to increase robustness to model inaccuracies.
    \textsc{SoftAE} uses optimistic, augmenting the mean model with \review{an auxiliary uncertainty variable that selects} the most favourable plausible dynamics within the model's uncertainty set.
    
    Although all three active exploration methods are theoretically guaranteed to converge to the true dynamics under mild assumptions, we show in \format{Supplementary} \format{Fig.}~\ref{figS:2_ae_comparison} that \textsc{SoftAE} achieves the most consistent and highest returns across tasks. In particular, \textsc{SoftAE} demonstrates clear advantages on challenging generalization scenarios such as \textit{(ii)} reach far, \textit{(iv)} turn \& swim, and \textit{(vi)} swim away.
    Returns are averaged over $10$ seeds for tasks \textit{(i)} and \textit{(ii)}, and $5$ seeds for the remaining tasks, shaded regions indicate standard error of the mean.

    \review{To better understand when optimistic planning provides a practical advantage over the alternatives, we repeat the experiments on the soft continuum arm while penalizing actuation effort. 
    We augment the intrinsic exploration reward \breview{$r_n^{\mathrm{exp}}$} with a quadratic action cost,}
    \[
    \review{
    r^{\mathrm{exp\breview{,act}}}_n(\vs_t,\va_t)
    =
    \norm{\vsigma_{n-1}(\vs_t,\va_t)}
    -
    c_{\mathrm{act}}\norm{\va_t}_2^2,
    }
    \]
    \review{where $c_{\mathrm{act}} \geq 0$ trades off information gain against actuation effort.
    We sweep $c_{\mathrm{act}}$ over the values $\{0,\, 0.25,\, 0.5,\, 0.75,\, 1.0\}$, with full results shown in \format{Supplementary Fig.}~\ref{figS:3_action_cost}.
    Without an action penalty ($c_{\mathrm{act}} = 0$), all three methods perform comparably, with a small advantage for \textsc{SoftAE}.
    Under moderate penalties ($c_{\mathrm{act}} \in \{0.25, 0.5, 0.75\}$), however, a clear separation emerges: \textsc{SoftAE} maintains its penalty-free performance and remains consistent across these three cost levels, whereas \textsc{Mean-AE} and \textsc{PETS-AE} degrade substantially.
    This separation isolates the practical consequence of the methods' differing treatment of plausible dynamics. 
    When exploration must seek informative states while limiting actuation effort, planning with the mean model or optimizing expected performance over sampled dynamics tends to favor conservative, low-effort trajectories that forgo informative but costly excursions. 
    By contrast, \textsc{SoftAE} can still identify a plausible dynamics realization within the confidence set under which an informative trajectory remains worthwhile, and therefore continues to explore the large-deformation regimes that matter for the harder tasks.
    At the largest penalty ($c_{\mathrm{act}} = 1.0$), the average \textsc{SoftAE} return also declines as a subset of seeds collapses to the same conservative behavior as \textsc{Mean-AE} and \textsc{PETS-AE}, indicating that a sufficiently strong action penalty can suppress informative exploration even under optimistic planning. Returns are averaged over $20$ random seeds.}

    \subsubsection*{\textsc{SoftAE} with Model-Based Policy Optimization}

        To investigate the integration of \textsc{SoftAE} with model-based policy optimization (MBPO) for obtaining closed-loop policies, we implement a proof-of-concept variant on the soft continuum arm environment. 
        In all cases, policies are trained with Soft Actor-Critic (SAC, c.f.~\cite{s-haarnoja2018sac}) and we use the implementation from~\cite{s-sukhija2024maxinforl}.
        Training uses sampled collected states augmented with simulated rollouts from the learned dynamics model, following standard MBPO protocols~\cite{s-hafner2019dream, s-janner2019trust, s-iten2025scalable}. 

        We compare our \textsc{SoftAE-MBPO} against three baselines: \textsc{Random}, which collects data from uncontrolled actuation without exploration or task guidance, and two standard task-oriented model-based policy optimization baselines, \textsc{MBPO} (using the mean model for simulated rollouts) and \textsc{MBPO-TS} (using an uncertainty-based trajectory sampling scheme).

        % We emphasize that these should be distinguished from the task-agnostic exploration baselines (\textsc{PETS-AE} and \textsc{Mean-AE}) in the main text, which guide exploration with uncertainty-based instead of task-specific rewards.
        
        As shown in \format{Supplementary} Fig.~\ref{figS:4_mbpo}, \textsc{SoftAE-MBPO} achieves performance on par with \textsc{MBPO} and \textsc{MBPO-TS} on their training task \textit{(i)}, but substantially outperforms them on previously unseen downstream tasks. 
        In contrast, the \textsc{Random} baseline lags behind across all tasks, reflecting the inefficiency of unguided data collection. The generalization of \textsc{SoftAE-MBPO} arises from its active exploration strategy, which prioritizes underexplored regions rather than overfitting to task-specific rewards.
        A known limitation, however, is the instability of policy optimization, since inaccuracies in the learned dynamics may mislead the policy early in training. 
        By contrast, trajectory optimization used in our main experiments re-optimizes open-loop action sequences at each time step, making it less sensitive to such bias. 
        Addressing this instability would require frequent retraining of the policy after each model update, which is computationally more expensive.

    \subsubsection*{\review{Comparison with Model-Free Reinforcement Learning on Real Hardware}}
    \label{secS:comparison_model_free}
    \review{\breview{Our framework demonstrates strong results with model-based trajectory planning and policy optimization. We next investigate whether similar performance can be achieved with model-free alternatives} with real-world interaction, without relying on a learned dynamics model.
    To examine this, we compare the real-world results in \format{Fig.}~\ref{fig:4_real_sopra}C against an online model-free reinforcement learning baseline, Soft Actor-Critic (SAC~\cite{s-haarnoja2018sac}).}

     \review{SAC here is the same algorithm \breview{and implementation} employed for policy training in our model-based variant, but trained directly on real environment transitions without any augmentation from synthetic rollouts of a learned dynamics model. }
     \breview{This model-free baseline is trained from scratch, without warm-up or pretraining of the actor or the critic.}
     \breview{Its hyperparameters are listed in Supplementary Table~\ref{tabS:5_sac_mf_params} and follow the standard SAC configuration~\cite{s-haarnoja2018sac,s-haarnoja2018soft}.} 
     \breview{We keep the same SoPrA reach-target reward formulation detailed in Supplementary Table~\ref{tabS:1_rewards}.}
     \breview{When trained on a single fixed target as in task \textit{(ix)}, SAC reaches a return of $17.7$ after 20 episodes ($4{,}000$ transitions, single seed\breview{; dashed curve in Supplementary Fig.~\ref{figS:5_sac}}), comparable to the task-specific model-based baseline \textsc{H-UCRL}.
     } 

    \breview{In contrast, for goal-conditioned reaching, a new target is sampled across the workspace at the start of each 200-step training episode.
    Learning this task requires larger sample sizes to obtain an accurate goal-conditioned value function estimate, which demands sufficient coverage under different target specifications.}
    \breview{The policy is then evaluated} \review{on the same two downstream reaching targets as in tasks \textit{(ix)} and \textit{(x)}} \breview{every 10 training episodes for one 20-step episode each}.
    \breview{We report the performance comparison with 3 random seeds in Supplementary Fig.~\ref{figS:5_sac}.}
    \breview{In a total of 200 training episodes, i.e.\ $200\times200=40{,}000$ real-world transitions, or 12\,h wall-clock time, SAC does not improve its return on the evaluation targets even with twice as much data as used in the model-based methods (\textsc{SoftAE}, \textsc{Random}, and \textsc{H-UCRL}).}
    \breview{With a limited online data stream, the amount and diversity of data are constrained and the replay buffer often reflects the robot’s own narrow visitation distribution. If the robot repeatedly visits only a relatively small region of the state--action space, the critic may have poor estimates elsewhere. Errors in the critic can propagate into the policy. The robot can subsequently move into regions where the critic has even less reliable data, producing a degradation. }
    
    \review{This gap \breview{is consistent with} a \breview{well-documented} challenge of online model-free methods in robotics: they require large amounts of real-world interaction to learn \breview{informative value functions or obtain accurate policy gradient estimates that eventually lead to performant control. Scaling such data} is prohibitively expensive on physical robots where each episode involves slow actuation, mechanical wear, and limited daily throughput.
    \breview{To our knowledge, few studies have trained model-free policies online on physical soft robots~\cite{s-morimoto2021elfnet,s-morimoto2022characterization,s-marquez2023hardware,s-avtges2025realtime}, some limited to quasi-static motion~\cite{s-marquez2023hardware}, while most model-free reinforcement learning for soft arms is instead trained in simulated environments where data collection can be parallelized~\cite{s-Naughton2021,s-jitosho2023reinforcement,s-satheeshbabu2019openloop,s-satheeshbabu2020continuous}. 
    On the same simulated Elastica arm used in our study, model-free policies were trained for 5--10 million transitions on target reaching and tracking tasks~\cite{s-Naughton2021}. On hardware, the smallest interaction budget we are aware of for learning a goal-conditioned task with model-free reinforcement learning is $100{,}000$ transitions~\cite{s-morimoto2021elfnet,s-morimoto2022characterization,s-avtges2025realtime}. \breview{Moreover}, hardware runs may not always converge~\cite{s-marquez2023hardware,s-avtges2025realtime} and \breview{can} show lower accuracy than model-based alternatives~\cite{s-avtges2025realtime}.
    }
    Model-based approaches such as \textsc{SoftAE} mitigate this cost by learning a reusable dynamics model that supports zero-shot planning across multiple downstream tasks without additional real-world data collection.
    This result reinforces the value of task-agnostic, uncertainty-driven exploration as a practical foundation for control on real-world soft robotic systems.}

    \subsubsection*{Environment Details}
    \label{secS:environment_details}
    We provide detailed descriptions of the state and action spaces for the three simulated soft robotic environments and one real-world setup used in our experiments: the soft continuum arm, the deformable fish in fluid, the musculoskeletal (MSK) leg, and the SoPrA soft continuum arm. 
    These systems differ in morphology, actuation, and task structure, resulting in varying observation state and control action dimensionalities.

    \format{Supplementary} Table~\ref{tabS:1_arm_state_action} outlines the state and action decomposition for the simulated soft continuum arm, which is discretized into either 5 or 11 elements along its body, yielding 58D and 136D state spaces, respectively. 
    The state includes spatially distributed position, linear velocity, orientation, and angular velocity entries, while the action space consists of torques applied in the normal and binormal directions at 6 control points. 

    \format{Supplementary} Table~\ref{tabS:1_fish_state_action} summarizes the state and action for the articulated fish with deformable skin, as used in tasks~\textit{(iii)} to~\textit{(vi)}. 
    The 15D state space applies to the simpler forward and U-turn swimming tasks~\textit{(iii)} and~\textit{(iv)}, while the 18D version includes an additional target-relative position vector for goal-reaching or avoiding tasks~\textit{(v)} and~\textit{(vi)}. 
    The fish is actuated by four internal joint torques.

    \format{Supplementary} Table~\ref{tabS:1_leg_state_action} describes the musculoskeletal (MSK) leg environment, which is controlled via a combination of a DC motor at the boom joint and four voltage inputs to antagonistic electrohydraulic muscles. 
    The state space captures the boom’s full orientation and angular velocity, along with the angles and velocities of the leg's two joints.
    
    Finally, \format{Supplementary} Table~\ref{tabS:1_sopra_state_action} presents the real-world soft continuum SoPrA arm environment. 
    The system is tracked using a motion capture software (Qualisys Track Manager) and actuated by six differential pressure commands, each applied to one of the arm’s air chambers.
    The state space consists of the position, linear velocity, orientation, and angular velocity of both the midpoint and tip motion-capture rings relative to the arm base ring (\format{Fig.}~\ref{fig:4_real_sopra}), along with the six previous step control commands, represented as normalized absolute pressures.
    
    \format{Supplementary} Table~\ref{tabS:1_rewards} details the reward functions used for each downstream task across the different environments, designed to align with the specific objectives of each task.
    
    \subsubsection*{Experiment Details}
    \label{secS:experiment_details}
        The hyperparameters used for our experiments are presented here.
        We train the dynamics model after each episode of data collection.
        For training, we fix the number of epochs to determine the number of gradient steps.
        The hyperparameters for dynamics model training, iCEM optimizer, and SAC policy training are presented in \format{Supplementary} \cref{tabS:2_agent_params,tabS:3_icem_params,tabS:4_sac_params}, respectively. The SAC hyperparameters correspond to the model-based policy training experiments on the simulated soft continuum arm, as reported in \format{Supplementary} Fig.~\ref{figS:4_mbpo}.

    \subsubsection*{\review{Computational Resources}}
    \review{Simulation experiments were conducted on a high-performance computing cluster, with each run allocated a single NVIDIA GeForce RTX 2080 Ti GPU (11 GB memory) and multi-core CPU resources. 
    Real-world experiments on the SoPrA arm were conducted on a workstation equipped with a single NVIDIA GeForce RTX 3090 GPU (24 GB memory) and a 16-core / 24-thread CPU. 
    Approximate wall-clock times for a single run were 1.3 hours for the soft continuum arm (100 episodes), 7.5 hours for the articulated fish in fluid (200 episodes), 21 hours for the musculoskeletal leg (250 episodes), and 6.5 hours for the real-world SoPrA arm (100 episodes). 
    Wall-clock times for the fish and musculoskeletal leg are dominated by the cost of the underlying physics simulators rather than by the exploration or model-learning components.
    Reported results for each method aggregate over 10 random seeds for tasks \textit{(i--ii)}, 5 seeds for tasks \textit{(iii--viii)}, and 3 seeds for the real-world tasks \textit{(ix--x)}.}

\FloatBarrier
\clearpage  

\begin{figure}[ht]
    \centering
    \includegraphics[width = 0.8\textwidth]{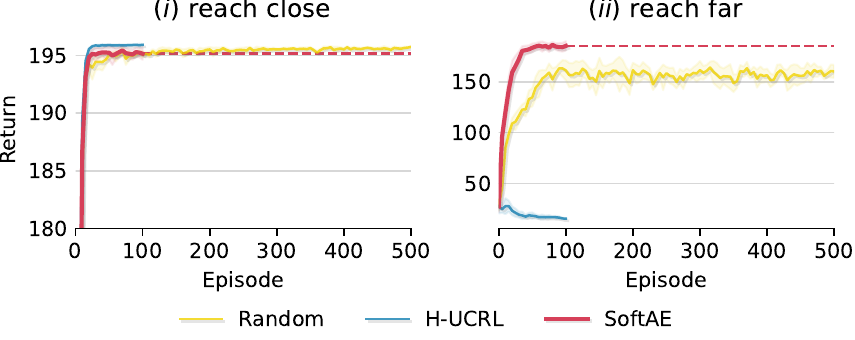}
    \caption{\review{\textbf{Random exploration with extended interaction budget on the simulated soft continuum arm.} 
    Return plots for the two downstream reaching tasks: task \textit{(i)} reach close, and task \textit{(ii)} reach far. \textsc{Random} is granted 500 exploration episodes, five times the budget given to \textsc{SoftAE} and \textsc{H-UCRL} (100 episodes each, identical to \format{Fig.}~\ref{fig:2_sim_tasks}A). 
    On the easier task \textit{(i)}, \textsc{Random} surpasses \textsc{SoftAE}'s performance at 100 exploration episodes and approaches the performance of \textsc{H-UCRL}, though without fully matching it. 
    On the harder task \textit{(ii)}, \textsc{Random} plateaus well below \textsc{SoftAE} even with the extended budget. Returns are averaged over 10 random seeds.}}
    \label{figS:1_random_extended}
\end{figure}

\FloatBarrier
\clearpage

\begin{figure}[ht]
    \centering
    \includegraphics[width = \textwidth]{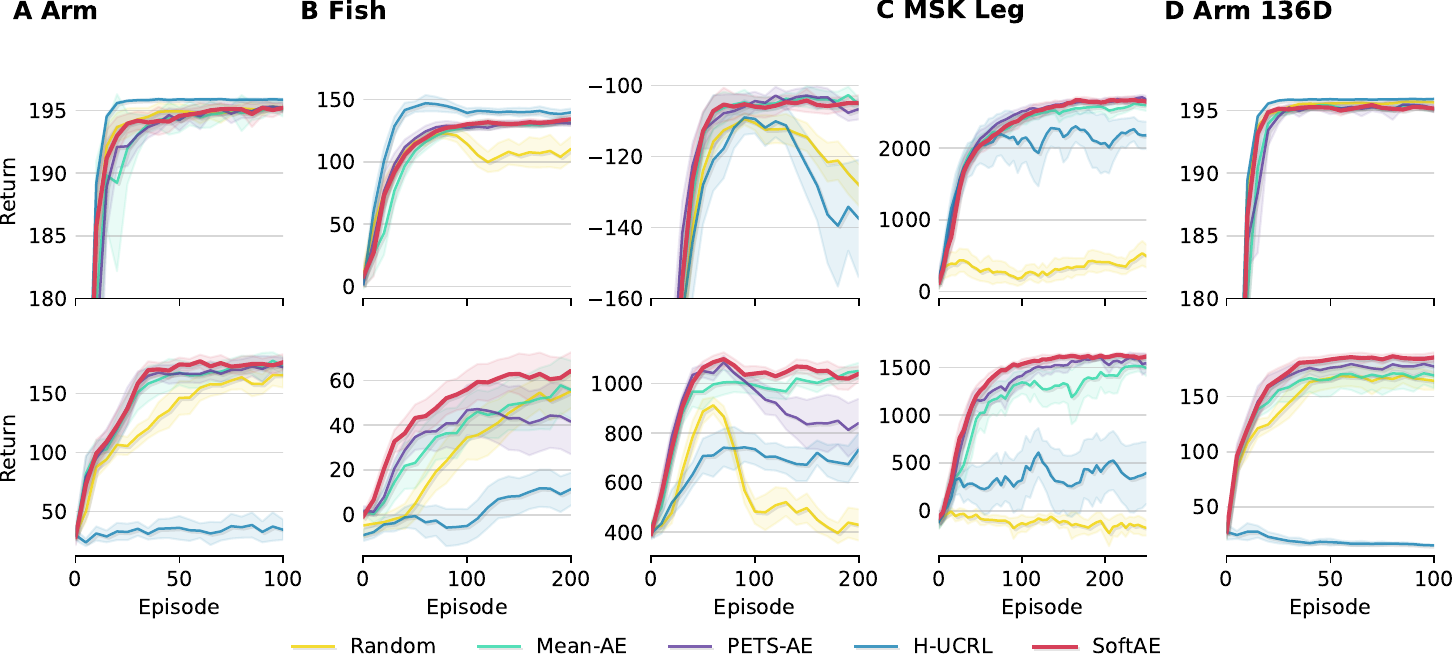}
    \caption{\textbf{Comparison of different uncertainty-driven active exploration strategies.}  
    Same as return plots in \format{Fig.}~\ref{fig:2_sim_tasks} but with two additional active exploration baselines: \textsc{Mean-AE} and \textsc{PETS-AE}.  
    \textsc{Mean-AE} plans trajectories using the mean model $\vmu_n$,  
    \textsc{PETS-AE} uses trajectory sampling from the model posterior to improve robustness to model inaccuracies.
    \review{Returns are averaged over 10 random seeds for tasks \textit{(i)} and \textit{(ii)}, and 5 seeds for the remaining tasks.}}
    \label{figS:2_ae_comparison}
\end{figure}

\FloatBarrier
\clearpage

\begin{figure}[ht]
    \centering
    \includegraphics[width = \textwidth]{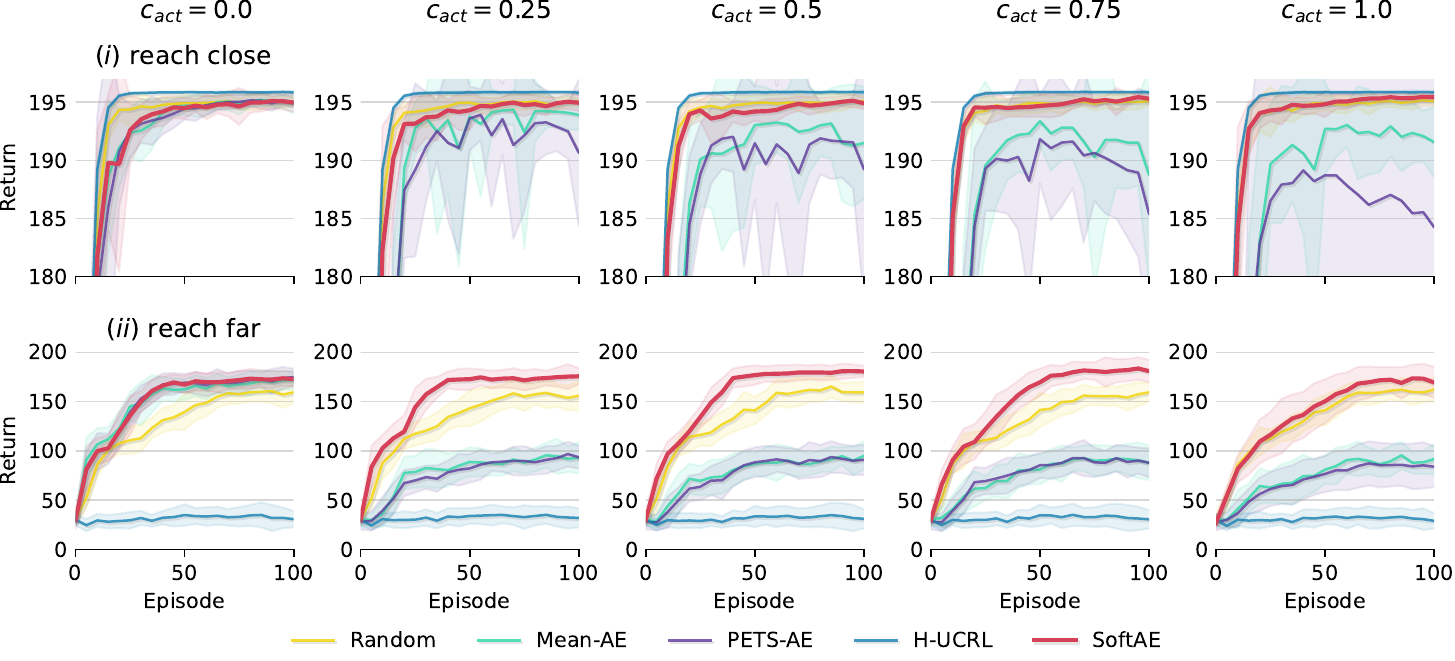}
    \caption{\review{\textbf{Effect of actuation cost on uncertainty-driven active exploration in the simulated soft continuum arm.}
    Same as the return plots in \format{Supplementary} \format{Fig.}~\ref{figS:2_ae_comparison}A, but with the intrinsic exploration reward augmented by an actuation penalty, $r_n^{\mathrm{exp,act}}(\vs_t, \va_t) = \norm{\vsigma_{n-1}(\vs_t, \va_t)} - c_{\mathrm{act}} \norm{\va_t}_2^2$. 
    Rows show downstream returns for tasks \textit{(i)} reach close (top) and \textit{(ii)} reach far (bottom); columns sweep $c_{\mathrm{act}} \in \{0.0, 0.25, 0.5, 0.75, 1.0\}$ (left to right). 
    Without a penalty, all three active exploration methods perform comparably. 
    Under moderate penalties ($c_{\mathrm{act}} \in \{0.25, 0.5, 0.75\}$), \textsc{SoftAE} maintains its penalty-free performance while \textsc{Mean-AE} and \textsc{PETS-AE} degrade heavily, as planning with the mean or sampled dynamics favors conservative trajectories. 
    At $c_{\mathrm{act}} = 1.0$, average \textsc{SoftAE} performance declines as a subset of seeds yields conservative trajectories similar to \textsc{Mean-AE} and \textsc{PETS-AE}, indicating that a sufficiently strong penalty can suppress informative exploration even under optimistic planning. Returns are averaged over 20 random seeds.}
    }
    \label{figS:3_action_cost}
\end{figure}

\FloatBarrier
\clearpage

\begin{figure}[ht]
    \centering
    \includegraphics[width = 0.6\textwidth]{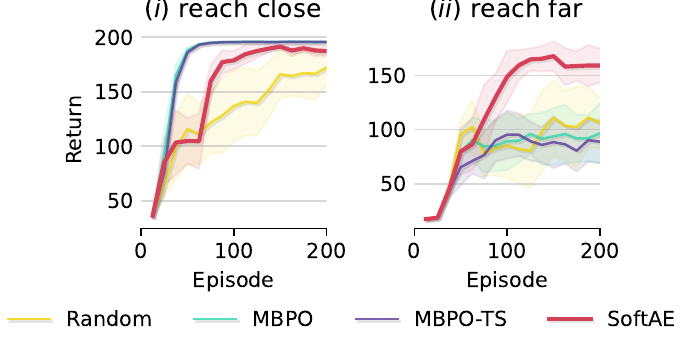}
    \caption{\textbf{Return comparison of policy optimization in the soft continuum arm simulation.}
    All methods use a learned dynamics model to train policies with SAC. Returns are averaged over 10 random seeds. \textsc{SoftAE-MBPO} employs uncertainty-based rewards for active exploration, while \textsc{MBPO} and \textsc{MBPO-TS} baselines are trained on task \textit{(i)}. The random baseline collects trajectories from random actuation. \textsc{SoftAE-MBPO} achieves performance comparable to task-specific baselines on the training task, but substantially outperforms them on unseen downstream tasks, highlighting the benefit of active exploration for generalization.}
    \label{figS:4_mbpo}
\end{figure}

\FloatBarrier
\clearpage

\begin{figure}[ht]
    \centering
    \includegraphics[width = 0.8\textwidth]{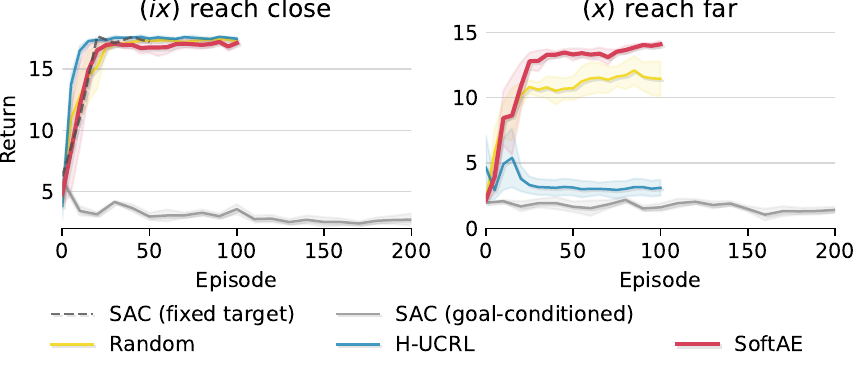}
    \caption{\review{\textbf{Comparison with model-free reinforcement learning on the real-world SoPrA arm.}
    % Same as return plots in \format{Fig.}~\ref{fig:4_real_sopra}C but with one additional baseline: Soft Actor-Critic (\textsc{SAC}).
    Same as return plots in \format{Fig.}~\ref{fig:4_real_sopra}C but with the additional model-free baseline: Soft Actor-Critic (\textsc{SAC}).
    \textsc{SoftAE}, \textsc{Random}, and \textsc{H-UCRL} use model-based planning with iCEM trajectory optimization on a learned dynamics model, 
    % while \textsc{SAC} is a model-free baseline trained directly on reaching randomized targets through online interaction with the physical arm, without any learned dynamics model. \textsc{SAC} fails to make meaningful progress within 200 episodes on both tasks, reflecting the high sample cost of model-free learning on real soft robotic hardware. Returns are averaged over 3 random seeds.
    % }}
    while \textsc{SAC} is a model-free baseline trained directly through online interaction with the physical arm, without any learned dynamics model. 
    \textsc{SAC} (fixed target) is trained on the single fixed target of task \textit{(ix)} only (dashed), while \textsc{SAC} (goal-conditioned) is trained on reaching randomized targets and evaluated on the fixed targets of tasks \textit{(ix)} and \textit{(x)}.
    Goal-conditioned \textsc{SAC} fails to make meaningful progress within 200 episodes on both tasks, reflecting the high sample cost of model-free learning on real soft robotic hardware. 
    Returns are averaged over 3 random seeds, with shaded regions indicating the standard error of the mean, except for \textsc{SAC} (fixed target), which is a single run.
    }}
    \label{figS:5_sac}
\end{figure}

\FloatBarrier
\clearpage

    \begin{table}[ht]
        \centering
        \caption{\textbf{State and action space decomposition of the soft continuum arm.} 
        This table details the individual state and action entries corresponding to the two observation discretization levels described in Table~\ref{tab:1_sim_env_summary}. 
        State entries include element positions, velocities, orientations, and angular velocities, while action entries correspond to applied torques at six equidistant points along the arm. 
        Dimensions are provided for both the 5-observed-point (58D) and 11-observed-point (136D) configurations.}
        \label{tabS:1_arm_state_action}
        \begin{tabular}{@{}lccc@{}}
        \toprule
        \textbf{State Entry} & \textbf{Symbol} & \textbf{Dimensions (58D)} & \textbf{Dimensions (136D)} \\ \midrule
        element positions           & $\mathbf{p}$              & $0{:}15$    & $0{:}33$    \\
        element linear velocities   & $\dot{\mathbf{p}}$        & $15{:}30$   & $33{:}66$   \\
        element orientations        & $\mathbf{R}$              & $30{:}46$   & $66{:}106$  \\
        element angular velocities  & $\boldsymbol{\omega}$     & $46{:}58$   & $106{:}136$ \\ \midrule
        \textbf{Action Entry} & \textbf{Symbol} & \textbf{Dimensions} \\ \midrule   
        normal torques              & $\boldsymbol{u}_n$     & $0{:}6$     \\
        binormal torques            & $\boldsymbol{u}_b$     & $6{:}12$    \\ \bottomrule
        \end{tabular}%
    \end{table}

\FloatBarrier
\clearpage

    \begin{table}[ht]
        \centering
        \caption{\textbf{State and action space decomposition of the articulated fish with deformable skin.} 
        This table follows the same format as \format{Supplementary} Table~\ref{tabS:1_arm_state_action}. 
        Dimensions are listed for task~\textit{(iii--iv)} with a 15D state space and task~\textit{(v--vi)} with an 18D state space, respectively.}
        \label{tabS:1_fish_state_action}
        \begin{tabular}{@{}lccc@{}}
        \toprule
        \textbf{State Entry} & \textbf{Symbol} & \textbf{Dimensions (15D)} & \textbf{Dimensions (18D)} \\ \midrule
        relative position of target     & $ \mathbf{p}_{\mathrm{target}} - \mathbf{p}$    & -     & $0{:}3$ \\
        base linear velocity            & $\mathbf{v}$          & $0{:}3$     & $3{:}6$     \\
        head orientation                & $\theta$              & $3$       & $6$       \\
        head angular velocity           & $\omega$              & $4$       & $7$       \\ 
        joint positions                 & $\mathbf{q}$          & $5{:}10$    & $8{:}13$    \\ 
        joint velocities                & $\dot{\mathbf{q}}$    & $10{:}15$   & $13{:}18$   \\ \midrule
        \textbf{Action Entry} & \textbf{Symbol} & \textbf{Dimensions} \\ \midrule   
        joint torques              & $\boldsymbol{u}$     & $0{:}4$     \\ \bottomrule
        \end{tabular}%
    \end{table}

\FloatBarrier
\clearpage

    \begin{table}[ht]
        \centering
        \caption{\textbf{State and action space decomposition for the musculoskeletal (MSK) leg with hybrid actuation.}
        This table follows the same format as \format{Supplementary} Table~\ref{tabS:1_arm_state_action}.}
        \label{tabS:1_leg_state_action}
        \begin{tabular}{@{}lcc@{}}
        \toprule
        \textbf{State Entry} & \textbf{Symbol} & \textbf{Dimensions} \\ \midrule
        boom yaw                & $\psi$                & $0$   \\
        boom pitch              & $\theta$              & $1$   \\
        boom roll               & $\phi$                & $2$   \\
        leg joint positions     & $\mathbf{q}$          & $3{:}5$ \\ 
        boom angular velocities & $\boldsymbol{\omega}$ & $5{:}8$ \\
        leg joint velocities    & $\dot{\mathbf{q}}$    & $8{:}10$\\ \midrule
        \textbf{Action Entry} & \textbf{Symbol} & \textbf{Dimensions} \\ \midrule   
        DC motor position target    & $u$   & $0$    \\
        muscle voltages             & $V$   & $1{:}5$ \\ \bottomrule
        \end{tabular}%
    \end{table}

\FloatBarrier
\clearpage

    \begin{table}[ht]
        \centering
        \caption{\textbf{State and action space decomposition for the real-world pneumatically actuated soft arm (SoPrA).}
        This table follows the same format as \format{Supplementary} Table~\ref{tabS:1_arm_state_action}.}
        \label{tabS:1_sopra_state_action}
        \begin{tabular}{@{}lcc@{}}
        \toprule
        \textbf{State Entry} & \textbf{Symbol} & \textbf{Dimensions} \\ \midrule
        tip capture ring position       & $\mathbf{p}$  & $0{:}3$   \\
        tip capture ring orientation    & $\mathbf{R}$  & $3{:}7$   \\
        midpoint capture ring position    & $\mathbf{p}_{\mathrm{mid}}$     & $7{:}10$   \\
        midpoint capture ring orientation & $\mathbf{R}_{\mathrm{mid}}$     & $10{:}14$   \\
        tip capture ring linear velocity & $\dot{\mathbf{p}}$           & $14{:}17$   \\
        tip capture ring angular velocity & $\boldsymbol{\omega}$       & $17{:}20$   \\
        midpoint capture ring linear velocity & $\dot{\mathbf{p}}_{\mathrm{mid}}$     & $20{:}23$   \\
        midpoint capture ring angular velocity  & $\boldsymbol{\omega}_{\mathrm{mid}}$  & $23{:}26$   \\
        previous commands (normalized absolute pressures)    & $\tilde{P}_{\mathrm{prev}}$          & $26{:}32$ \\  \midrule
        \textbf{Action Entry} & \textbf{Symbol} & \textbf{Dimensions} \\ \midrule   
        delta pressures    & $\Delta P$   & $0{:}6$    \\ \bottomrule
        \end{tabular}%
    \end{table}

\FloatBarrier
\clearpage

    \begin{table}[ht]
        \centering
        \caption{\textbf{Downstream task rewards for simulated and real-world environments.} 
        This table summarizes the reward definitions for all simulated environments described in Table~\ref{tab:1_sim_env_summary} as well as for the real-world SoPrA platform. 
        Each task reward is formulated as a function of position, linear velocity, or angular displacement, and follows the notation introduced in \format{Supplementary} Tables~\ref{tabS:1_arm_state_action}--\ref{tabS:1_sopra_state_action}, depending on the environment and objective.}
        \label{tabS:1_rewards}
        \begin{tabular}{@{}llc@{}}
            \toprule
            \textbf{Environment}        & \textbf{Task} 
                                        & \textbf{Reward} $r_t$ \\ \midrule
            \multirow{1}{*}{Arm}        & \textit{(i) - (ii)} reach target                 
                                        & $g(\left\| \mathbf{p}_t - \mathbf{p}_{\mathrm{target}} \right\|_2)$\footnotemark[1] \\ 
            \midrule
            \multirow{4}{*}{Fish}       & \textit{(iii)} swim forward along +x direction   
                                        & $v_{+x,t}$ \\
                                        & \textit{(iv)} turn and swim along -x direction   
                                        & $v_{-x,t}$     \\
                                        & \textit{(v)} swim to target                      
                                        & $-\left\| \mathbf{p}_t - \mathbf{p}_{\mathrm{target}} \right\|_2$ \\
                                        & \textit{(vi)} swim away from target              
                                        & $\left\| \mathbf{p}_t - \mathbf{p}_{\mathrm{target}} \right\|_2$  \\ 
            \midrule
            \multirow{2}{*}{MSK Leg}    & \textit{(vii)} hop counterclockwise  
                                        & $\dot{\phi_t}$                      \\
                                        & \textit{(viii)} hop clockwise
                                        & $-\dot{\phi_t}$         \\ 
            \midrule
            \multirow{1}{*}{SoPrA}      & \textit{(ix)-(x)} reach target 
                                        & $g(\left\| \mathbf{p}_t - \mathbf{p}_{\mathrm{target}} \right\|_2)$\footnotemark[1]                      \\ 
            \bottomrule
            \end{tabular}%
        \end{table}
        \footnotetext[1]{For arm reach tasks, we employ a shaped long-tail reward function that maps the distance to target into the range $(0, 1]$. Specifically,  $g(\left\| \mathbf{p}_t - \mathbf{p}_{\mathrm{target}} \right\|_2) = 1$ when $\left\| \mathbf{p}_t - \mathbf{p}_{\mathrm{target}} \right\|_2 \leq 5 \times 10^{-3}$.}

\FloatBarrier
\clearpage

        \begin{table}[ht]
        \centering
        \caption{\textbf{Hyperparameters used in the experiments.} 
        This table lists the dynamics model training hyperparameters for all soft robotic environments. 
        The reported values cover exploration and task horizons, dynamics model network architecture and ensemble size, and training schedule.}
        \label{tabS:2_agent_params}
        \begin{tabular}{@{}lcccc@{}}
            \toprule
            \textbf{Hyperparameters}                  & \textbf{Arm}   & \textbf{Fish}  & \textbf{MSK Leg} & \textbf{SoPrA} \\ \midrule
            Exploration horizon              & $200$   & $200$   & $500$     & $200$   \\
            Downstream task horizon          & $200$   & $200$   & $500$     & $20$    \\
            Hidden layers                    & $4$     & $4$     & $4$       & $4$     \\
            Neurons per layer               & $256$   & $256$   & $256$     & $256$   \\
            Number of ensembles              & $5$     & $5$     & $5$       & $5$     \\
            Batch size                       & $64$    & $64$    & $64$      & $64$    \\
            Learning rate         
            & $5\times 10^{-5}$  
            & $5\times 10^{-5}$ 
            & $5\times 10^{-5}$  
            & $5\times 10^{-5}$      \\
            Number of epochs                 & $50$    & $50$    & $50$      & $50$    \\
            Maximum number of gradient steps & $5,000$ & $7,500$ & $5,000$   & $5,000$ \\
            $\beta$                        & $2.0$ & $2.0$ & $2.0$   & $2.0$ \\ \bottomrule
            \end{tabular}%
        \end{table}

\FloatBarrier
\clearpage    
        
        \begin{table}[ht]
        \centering
        \caption{\textbf{Hyperparameters of the iCEM optimizer.} 
        This table reports the optimizer settings used in this work, including the number of samples, horizon length, elite-set size, colored-noise exponent, number of particles, CEM iterations, and the fraction of elites reused. 
        Values are shown separately for each environment and task group.}
        \label{tabS:3_icem_params}
        \begin{tabular}{@{}lccccc@{}}
        \toprule
            \textbf{Hyperparameters} & 
            \textbf{Arm} & 
            \textbf{Fish - (iii) \& (iv)} & 
            \textbf{Fish - (v) \& (vi)} & 
            \textbf{MSK Leg} & \textbf{SoPrA} \\ \midrule
            Number of samples $P$        & $200$  & $200$  & $200$  & $200$  & $200$ \\
            Horizon $H$                  & $10$   & $10$   & $100$  & $100$  & $5$   \\
            Size of elite-set $K$       & $20$   & $20$   & $20$   & $20$   & $20$  \\
            Colored-noise exponent    & $0.25$ & $0.25$ & $0.25$ & $0.25$ & $1.0$ \\
            Number of particles       & $10$   & $10$   & $10$   & $10$   & $10$  \\
            CEM-iterations            & $5$    & $5$    & $5$    & $5$    & $5$   \\
            Fraction of elites reused $\xi$ & $0.3$  & $0.3$  & $0.3$  & $0.3$  & $0.3$ \\ \bottomrule
        \end{tabular}
        \end{table}
        
\FloatBarrier
\clearpage

        \begin{table}[ht]
        \centering
        \caption{\textbf{Hyperparameters of the SAC algorithm in the model-based RL setting.} 
        This table lists the training hyperparameters used for the SAC agent, including exploration and task horizons, network architecture, ensemble size, batch size, learning rate, and imagined rollout horizon.}
        \label{tabS:4_sac_params}
        \begin{tabular}{@{}lc@{}}
        \toprule
            \textbf{Hyperparameters} & 
            \textbf{Arm} \\ \midrule

            Exploration horizon              & $200$\\
            Downstream task horizon          & $200$\\
            Hidden layers                    & $3$\\
            Neurons per layer               & $1024$\\
            Number of ensembles              & $5$ \\
            Batch size                       & $256$ \\
            Learning rate & $3\times 10^{-4}$\\
            Imagined rollout horizon $H$                  & $5$   \\
        \bottomrule
        \end{tabular}
        \end{table}
        
\FloatBarrier
\clearpage

\begin{table}[ht]
        \centering
        \caption{\breview{
        \textbf{Hyperparameters of the SAC algorithm in the online model-free RL setting.}
        This table lists the training hyperparameters used for the SAC baseline of Supplementary Fig.~\ref{figS:5_sac}. The code implementation is the same as in the model-based setting (Supplementary Table~\ref{tabS:4_sac_params}), but this model-free baseline is trained on real environment transitions only. 
        The reward is the SoPrA reach target reward of Supplementary Table~\ref{tabS:1_rewards}.
        }}
        \label{tabS:5_sac_mf_params}
        \begin{tabular}{@{}lc@{}}
        \toprule
            \textbf{Hyperparameters} &
            \textbf{SoPrA} \\ \midrule
            Episode horizon                          & $200$ \\
            Training episodes                        & $200$ \\
            Total environment transitions            & $40{,}000$ \\
            Random warm-up transitions               & $0$ \\
            Update-to-data ratio                     & $1$ \\
            Hidden layers                            & $2$ \\
            Neurons per layer                        & $256$ \\
            Batch size                               & $256$ \\
            Learning rate (actor, critic, $\alpha$)  & $3\times 10^{-4}$ \\
            Discount factor $\gamma$                 & $0.99$ \\
            Target smoothing coefficient $\tau$      & $0.005$ \\
            Entropy coefficient $\alpha$             & auto-tuned, initial $1.0$ \\
            Target entropy                           & $-6$ \\  %% 
            Replay buffer size                       & $10^{6}$ \\
            Evaluation                               & $1$ episode each on tasks \textit{(ix)} and \textit{(x)}\\            
            Evaluation frequency                     & $10$ \\
            Evaluation episode horizon               & $20$ \\
            Random seeds                             & $3$ \\
        \bottomrule
        \end{tabular}
        \end{table}

\FloatBarrier
\clearpage

\subsection*{Legends for Supplementary Movies 1 to 5}

\paragraph{Supplementary Movie 1:}
\textbf{Soft-robot Active Exploration (SoftAE).} This video gives an overview of \textsc{SoftAE}, a generalizable active exploration framework for efficient soft robotic dynamics learning and zero-shot control.

\url{https://www.youtube.com/watch?v=WfOc6DzHEow}
\\
\paragraph{Supplementary Movie 2:}
\textbf{Simulated soft continuum arm via Cosserat rods.}
This video first presents the environment setup, including the state and action space decomposition (see Supplementary Table~1), before demonstrating the downstream reaching tasks~\textit{(i)--(ii)}. 
Agent performance is compared across three exploration strategies: \textsc{Random}, \textsc{H-UCRL}, and our proposed \textsc{SoftAE}. 
The video further shows how downstream task performance improves steadily as more exploration episodes are collected with \textsc{SoftAE}.
% Simulated soft continuum arm via Cosserat rods. This video first presents the environment setup, including the state and action space decomposition (see Supplementary Table 1), before demonstrating the downstream reaching tasks (i)-(ii). Agent performance is compared across three exploration strategies: Random, H-UCRL, and our proposed SoftAE. The video further shows how downstream task performance improves steadily as more exploration episodes are collected with SoftAE.

\url{https://www.youtube.com/watch?v=-3oxO-yeP5g}
\\
\paragraph{Supplementary Movie 3:}
\textbf{Articulated fish with deformable skin in fluid simulation.}
This video follows the same structure as \format{Supplementary} Movie \format{2}, beginning with the environment setup which comprises both the state--action space decomposition (see \format{Supplementary} Table~2), and the particle-based fluid simulation, illustrating fluid dynamics and fluid--structure interaction.
It then demonstrates the downstream swimming tasks~\textit{(iii)--(vi)}, with performance compared across \textsc{Random}, \textsc{H-UCRL}, and the proposed \textsc{SoftAE}, as in \format{Supplementary} Movie \format{2}.
% Articulated fish with deformable skin in fluid simulation. This video follows the same structure as Supplementary Movie 2, beginning with the environment setup which comprises both the state-action space decomposition (see Supplementary Table 2), and the particle-based fluid simulation, illustrating fluid dynamics and fluid-structure interaction. It then demonstrates the downstream swimming tasks (iii)-(vi), with performance compared across Random, H-UCRL, and the proposed SoftAE, as in Supplementary Movie 2.

\url{https://www.youtube.com/watch?v=bGTE4vuSoVE}
\\
\paragraph{Supplementary Movie 4:}
\textbf{Simulated musculoskeletal leg powered by electrohydraulic muscles.}
This video begins by illustrating the electrohydraulic muscle actuation, followed by the system setup, and the state--action space decomposition (see \format{Supplementary} Table~3). 
It then demonstrates the downstream hopping tasks~\textit{(vii)--(viii)}, with performance compared across \textsc{Random}, \textsc{H-UCRL}, and the proposed \textsc{SoftAE}, as in Supplementary Movie 2.
% Simulated musculoskeletal leg powered by electrohydraulic muscles. This video begins by illustrating the electrohydraulic muscle actuation, followed by the system setup, and the state-action space decomposition (see Supplementary Table 3). It then demonstrates the downstream hopping tasks (vii)-(viii), with performance compared across Random, H-UCRL, and the proposed SoftAE, as in Supplementary Movie 2.

\url{https://www.youtube.com/watch?v=rwUSm3N94XE}
\\
\paragraph{Supplementary Movie 5:}
\textbf{Real-world pneumatically actuated SoPrA arm.}
This video presents the real-world SoPrA environment and task performance and is the animated equivalent of Fig.~4.
% Real-world pneumatically actuated SoPrA arm. This video presents the real-world SoPrA environment and task performance and is the animated equivalent of Fig.4.

\url{https://www.youtube.com/watch?v=l0UKPQhPSJ8}

\FloatBarrier
\clearpage

\renewcommand{\refname}{Supplementary References}

% \newpage
% \input{3-X-DescriptionAdditionalSupplementaryFiles}

\end{document}